\documentclass[letterpaper]{article} 
\usepackage[preprint]{aaai2027}
\usepackage[hyphens]{url}  
\usepackage{graphicx} 
\usepackage{natbib}  
\usepackage{caption} 
\usepackage{algorithm}
\usepackage{algorithmic}

\usepackage{amsmath}
\usepackage{amssymb}
\usepackage{amsfonts}

\usepackage{subcaption}
\usepackage{booktabs}
\usepackage{multirow}
\usepackage{makecell}
\usepackage{newfloat}
\usepackage{listings}
\DeclareCaptionStyle{ruled}{labelfont=normalfont,labelsep=colon,strut=off} 
\floatstyle{ruled}
\newfloat{listing}{tb}{lst}{}
\floatname{listing}{Listing}

\title{Drive by Hindsight and Foresight: Tool-Grounded Synergistic Reasoning over Hierarchical Memory for Autonomous Driving}

\author{
	Baojie Chen\textsuperscript{\rm 1}\corresponding,
	Zijun Jia\textsuperscript{\rm 1},
	Jing Zhong\textsuperscript{\rm 2}
}
\affiliations{
	\textsuperscript{\rm 1} Beihang University\\
	\textsuperscript{\rm 2} Tsinghua University
}

\begin{document}

\maketitle

\begin{abstract}
Vision-language models (VLMs) have shown promise for autonomous driving tasks, yet still suffer from hallucination, weak spatio-temporal perception, and limited generalization. Recent methods improve reasoning and decision-making capabilities through chain-of-thought explanations, retrieval-augmented generation or the static incorporation of external tool outputs. Although these mechanisms enrich the initial context, the model neither proactively perceives scene information nor accumulates experience after answering. To overcome these limitations, we present, to our knowledge, the first synergistic framework that tightly couples hierarchical memory with proactive tool invocation in a closed reasoning loop. Our contributions are threefold. \textbf{(i) Hierarchical Driving Memory}: a scene-level short-term memory maintains the dynamic scene state, and an evolving long-term memory retrieves reusable experience and tool strategies by similarity. \textbf{(ii) Memory-Tool Synergistic Reasoning Framework}: guided by the scene state and retrieved experience, the model adaptively invokes tools to refine its reasoning at inference time and consolidates reusable experience into a long-term memory pool offline. \textbf{(iii) Data Generation and Two-stage Training Pipeline}: verified memory-tool trajectories built by multi-step teacher rollout are used to train with stepwise supervised fine-tuning (SFT) and group relative policy optimization (GRPO). Our 7B model reaches an overall reasoning score of 80.03 and MCQ accuracy of 79.09\%  on DriveLMM-o1 benchmark, surpassing the strongest baseline by 7.74 MCQ points and generalizes strongly across
DriveMLLM and STRIDE-QA benchmarks. Notably, ablation and analysis studies validate the effectiveness of each component and further reveal the
complementary roles of hierarchical memory. Short-term memory strengthens
spatiotemporal understanding, improving STSBench accuracy by 24.2 points,
while offline long-term memory consolidation yields an additional
3.57-point MCQ gain with all model parameters frozen, demonstrating
continual self-evolution through accumulated driving experience.

\end{abstract}


\section{Introduction}
\label{sec:intro}

\begin{figure}[t]
\centering
\includegraphics[width=\columnwidth]{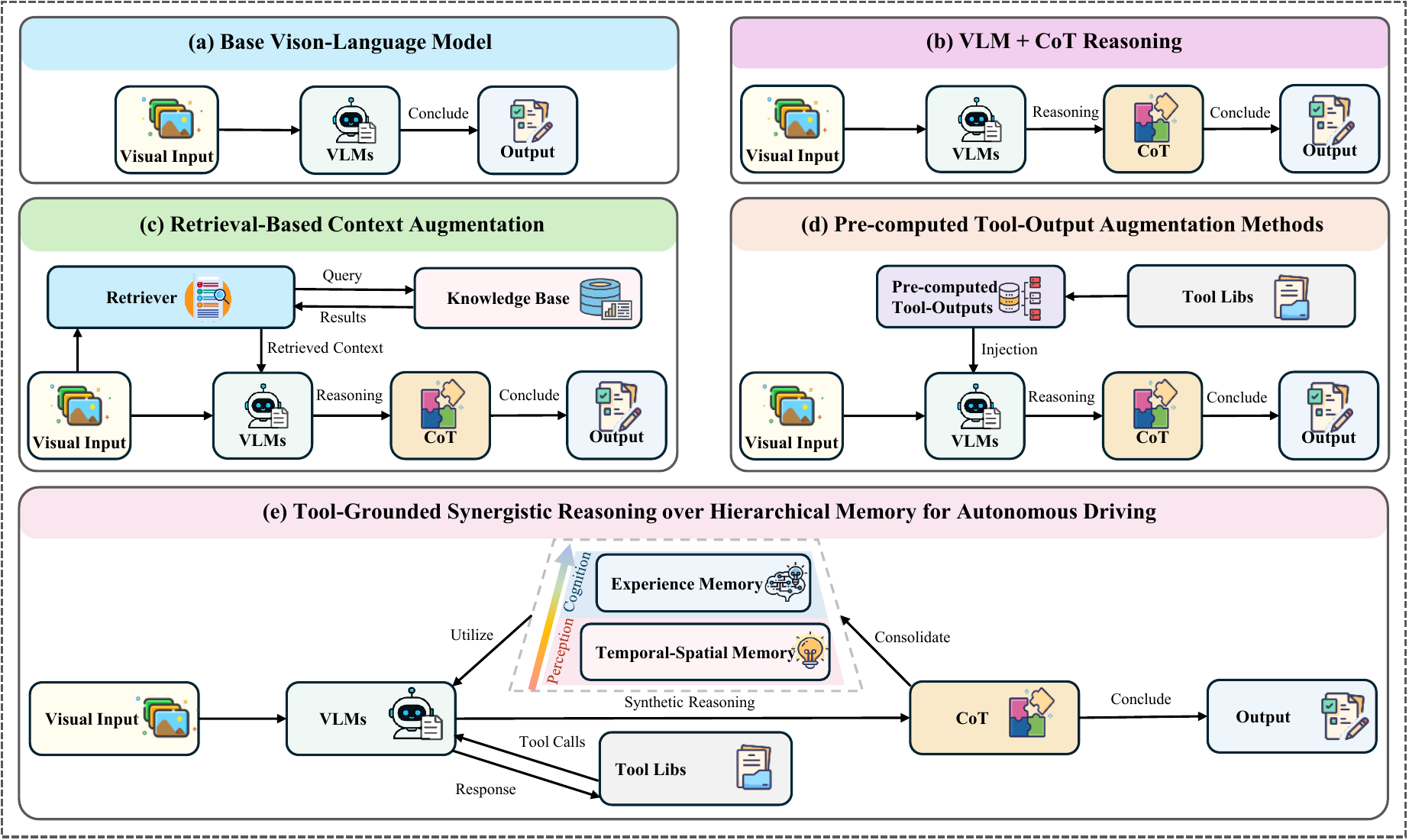}
\caption{Comparison of reasoning paradigms for autonomous driving:
		(a) A base VLM directly maps visual inputs to answers;
		(b) CoT unfolds direct prediction into guided intermediate reasoning steps; (c) RAG augments reasoning with retrieved external context;
		(d) Previous tool-augmented methods inject pre-computed tool outputs into
		single-pass reasoning; and (e) Our framework couples hierarchical memory
		with online tool interaction and offline experience consolidation.} 
\label{fig:contrast}
\end{figure}

Pretrained vision-language models have enabled their growing adoption in autonomous driving~\citep{tianDriveVLM2024,hwangEMMA2024,jiangSennaBridgingLarge2024a,fuORIONHolisticEndtoend2025,nvidiaAlpamayoR12025}, and are increasingly used for driving scene understanding, high-level behavior decisions, and decision explanation~\citep{jiangSennaBridgingLarge2024a,maoLanguageAgentAutonomous2024}. Among these applications, autonomous driving reasoning benchmarks provide a natural testbed for evaluating whether
VLMs can transform visual observations into interpretable scene
understanding and driving decisions.~\citep{marcuLingoQA2024,xieAreVLMsReady2025a,ishaqDriveLMMo1StepbystepReasoning2025}.

Existing efforts proceed largely along three routes: unfolding the observation into step-by-step explanations via chain-of-thought (CoT) for better transparency and interpretability~\citep{ishaqDriveLMMo1StepbystepReasoning2025,zengFutureSightDriveThinkingVisually2025}; splicing the outputs of driving tools into the prompt or context to supplement external observations~\citep{qianAgentThinkUnifiedFramework2025,zhengDriveAgentR12025}; or injecting knowledge via retrieval-augmented generation (RAG)~\citep{yuanRAGdriverGeneralisableDriving2026,yeSafeDriveRAGSafeAutonomous2025,wangRADRetrievalaugmentedDecisionmaking2025}. In all three, the model is either made to produce more elaborate explanations from templates or to passively receive more information, the answers it produces and the information it receives keep growing, while its understanding does not evolve accordingly. Human driving follows a different pattern: a driving decision is not a one-shot recognition of a static frame but a unified process of active perception, spatio-temporal understanding, and experience accumulation. A driver relies on \emph{foresight} to keep track of the motion trends of the ego vehicle and key objects over time, and on \emph{hindsight} to draw on transferable experience accumulated through long-term driving, judging which risks to prioritize and which actions to take in the current situation. Against this reference, what current methods lack becomes clear: no dynamic scene state that evolves with observations, no driving experience consolidated across scenes, and no active evidence-seeking driven by evidence gaps. This gap motivates the central question of this paper: \emph{Can a reasoning model for autonomous driving capture the key characteristics of human driving foresight and hindsight by integrating spatio-temporal awareness, proactive perception, and experience accumulation within a unified closed-loop reasoning framework?}

To this end, we propose a tool-grounded synergistic reasoning
framework over hierarchical memory. Scene-level short-term
memory maintains the current spatio-temporal state, while
cross-scene long-term memory retrieves experience as
transferable priors. Conditioned on hierarchical memory, the model
proactively invokes tools to acquire information. Across three driving benchmarks, our 7B model outperforms the strongest baseline and generalizes robustly without further training.
Controlled studies further
show that short-term memory improves STSBench accuracy by
$24.2$ points, while offline long-term memory consolidation
provides an additional $3.57$-point MCQ gain without updating
model parameters.
More broadly, these results point to a shift
in autonomous-driving VLM research: from scaling static
perception and isolated reasoning toward persistent,
interactive driving intelligence that can evolve with evidence
and experience.Our contributions are summarized as
follows:
\begin{itemize}
\item \textbf{(1) We develop a hierarchical driving memory} combing scene-level short-term memory for maintaining spatio-temporal states with cross-scene long-term memory for retrieving reusable driving experience and strategies.
\item \textbf{(2) We propose a memory-tool synergistic reasoning (MTSR) framework} which unifies foresight and hindsight by coupling adaptive tool use with hierarchical memory at inference time and consolidates verified experience into long-term memory offline.
\item \textbf{(3) We build a data-generation and two-stage post-training pipeline} that produces verified memory-tool trajectories through multi-step teacher rollouts and trains the model with stepwise supervised fine-tuning (SFT) and group relative policy optimization (GRPO).
\end{itemize}

\section{Related Work}
\label{sec:related}

\subsection{VLMs and Benchmarks for Autonomous Driving}

Vision-language models (VLMs) increasingly connect visual perception with
language-based reasoning and planning in autonomous driving, with systems such
as Senna, ORION, and NaviDriveVLM exploring different interfaces between high-level
reasoning and motion
planning~\citep{jiangSennaBridgingLarge2024a,fuORIONHolisticEndtoend2025,taoNaviDriveVLMDecouplingHighlevel2026}. In parallel, the benchmarks have progressed from perception-oriented and free-form question answering to risk assessment, explicit reasoning, and spatial-temporal understanding as well~\citep{simaDriveLM2024,qianNuScenesQA2024,marcuLingoQA2024,gaoNuRiskVisualQuestion2025,ishiharaSTRIDEQA2026,gholamiSpatialReasoning2025}.DriveMLLM evaluates spatial understanding~\cite{guoDriveMLLMBenchmark2024}, and STSBench emphasizes spatio-temporal scenario understanding~\citep{fruhwirthSTSBench2025}. These benchmarks, together with reliability studies~\citep{xieAreVLMsReady2025a,yuWaymoQA2025}, show persistent weaknesses in motion trends, object interactions, grounding, and safety-critical reasoning. 

\subsection{External Augmentation in Reasoning for Autonomous Driving}

Chain-of-thought (CoT) methods make intermediate driving judgments more explicit. DriveLMM-o1 supervises step-by-step reasoning for scene understanding~\citep{ishaqDriveLMMo1StepbystepReasoning2025}, FutureSightDrive introduces spatio-temporal visual CoT~\citep{zengFutureSightDriveThinkingVisually2025}, and reinforcement- and critic-based approaches improve reasoning for planning and safety assessment~\citep{zhangOmniDriveR1ReinforcementdrivenInterleaved2026,yangJudgeThenDrive2026,liuCritiqueDriveVLM2026}. External augmentation provides complementary information: SafeDriveRAG, RAG-Driver, and RAD retrieve knowledge or driving cases for safety reasoning, explanation, and decision making~\citep{yeSafeDriveRAGSafeAutonomous2025,yuanRAGdriverGeneralisableDriving2026,wangRADRetrievalaugmentedDecisionmaking2025}. AgentThink takes a step further through adding tool-outputs into context to improve performance and alleviate hallucinations~\citep{qianAgentThinkUnifiedFramework2025}. However, all the methods above still rely on a one-way, passive intake process rather than understanding that evolves with evidence, which motivates us to propose our framework.

\section{Methodology}
\label{sec:method}

\subsection{Overview}
\label{sec:overview}

\begin{figure*}[t]
\centering
\includegraphics[width=0.95\textwidth]{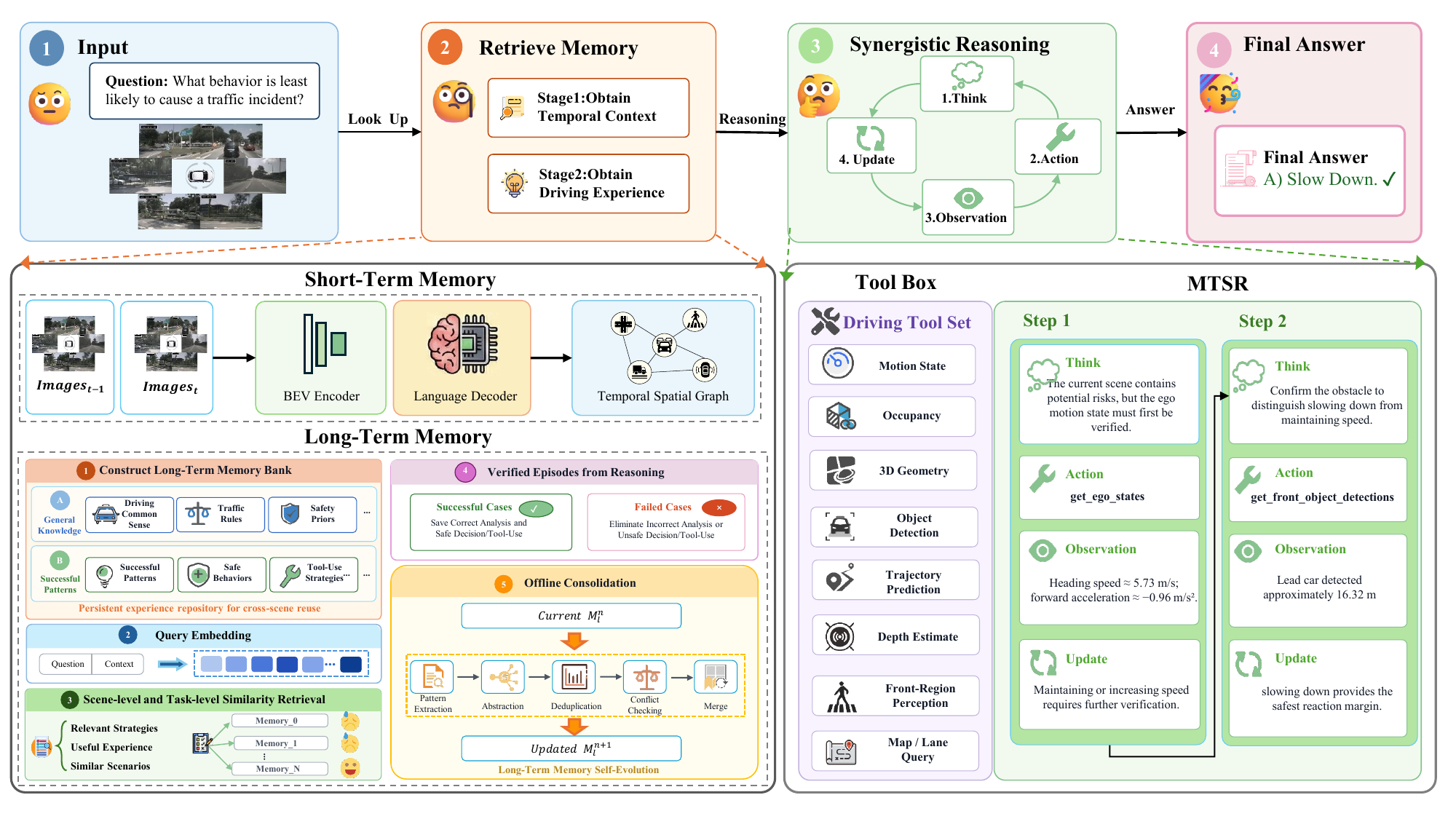}
\caption{Overview of our framework. Given a question and a multi-view image, the model conditions on a scene-level short-term memory $M_s$ and experience retrieved from a cross-scene long-term memory $M_l$, then performs Memory-Tool Synergistic Reasoning: it identifies the current evidence gap, invokes a driving tool, grounds the observation, and updates the scene belief, looping until it answers.}
\label{fig:overview}
\end{figure*}

As shown in Figure~\ref{fig:overview}, given a question $q$ and
the current six-view observation $\mathcal{I}_t=\{I_t^v\}_{v=1}^{6}$,
our framework conditions the vision-language policy $\pi_\theta$ on scene-level
short-term memory $\mathcal{M}_s$, the top-$K$ entries
$\mathcal{E}_K$ retrieved from cross-scene long-term memory
$\mathcal{M}_l$, and the driving tool set $\mathcal{T}$:
\begin{equation}
	y \sim \pi_\theta
	(q,\mathcal{I}_t,\mathcal{M}_s,\mathcal{E}_K,\mathcal{T}).
\end{equation}
Here, $\mathcal{M}_s$ represents the current spatio-temporal
state, whereas $\mathcal{E}_K$ provides reusable experience.
The policy iteratively identifies missing evidence, invokes
tools, accumulates the returned observations in its reasoning context, until it judges the evidence sufficient and answers.

\subsection{Hierarchical Driving Memory}
\label{sec:memory}

Our framework maintains a hierarchical driving memory consisting of $M_s$ and
$M_l$. The former provides a compact spatio-temporal representation of
the current scene, whereas the latter supplies reusable reasoning and tool-use experience from past scenes.

\subsubsection{Scene-Level Short-Term Memory.}

A single-timestamp multi-view observation provides rich appearance
information but cannot explicitly reveal how the ego vehicle and nearby
traffic participants evolve over time. Directly feeding multiple
six-view frames into the VLM would substantially increase the visual
context and require the language model to perform cross-view alignment
and temporal correspondence implicitly. We therefore construct $M_s$
from two adjacent six-view observations $\mathcal{I}_{t-1}$ and $\mathcal{I}_t$.

For each timestamp $\tau\in\{t-1,t\}$, a pretrained bird’s-eye view (BEV)
perception model~\cite{zhouHERMES2025} encodes the surround-view images into an
ego-centric BEV representation and predicts both a structured semantic
description and scene attributes:
\begin{equation}
    (d_\tau,n_\tau)=F_{\mathrm{BEV}}(\mathcal{I}_\tau),
	\label{eq:bev_parsing}
\end{equation}
where $d_\tau$ describes traffic participants, road elements, $n_\tau$ contains locations, qualitative relations and motion trends. Then we organize them into a scene graph:
\begin{equation}
	 M_s = \Phi_{\mathrm{graph}}(d_{t-1}, d_t, n_{t-1}, n_t) = (\mathcal{V}, \mathcal{R}),
	\label{eq:scene_graph}
\end{equation}
where $\mathcal{V}$ contains the ego vehicle, traffic participants, and
relevant road elements; $\mathcal{R}$ represents their spatial,
temporal, and interaction relations. Notably, the short-term memory is refreshed only when the input scene advances and remains a read-only scene prior within a trajectory.

\subsubsection{Cross-Scene Long-Term Memory.}
Reliable reasoning also requires hindsight. To maintain a persistent
long-term memory $M_l$ containing driving commonsense, traffic rules and reusable experience distilled from verified successful trajectories, we define each entry as 
$e_i=(c_i,r_i,p_i^{+},\pi_i^{\mathrm{tool}},w_i)$ , where $c_i$ describes the applicable scene and task conditions, $r_i$
summarizes relevant risk factors, $p_i^{+}$ denotes a reusable
successful reasoning pattern, $\pi_i^{\mathrm{tool}}$ records an
effective tool-use strategy, and $w_i$ denotes its retrieval utility.
For a question $q$, we jointly encode the question and $M_s$ into a query representation,
ranking each entry by semantic relevance and normalized utility:
\begin{equation}
	\begin{aligned}
		S(e_i\mid q,M_s)
		&=
		\operatorname{sim}
		\left(
		E_m([q;M_s]),E_m(e_i)
		\right)
		+\lambda\bar{w}_i,\\
		\mathcal{E}_K
		&=
		\operatorname{TopK}_{e_i\in M_l}
		S(e_i\mid q,M_s).
	\end{aligned}
\end{equation}
where $\bar{w}_i$ is the normalized utility and $\lambda$
controls its contribution. After filtering and deduplication,
$\mathcal{E}_K$ provides transferable reasoning and tool-use
priors rather than direct answer hints.

\begin{figure*}[t]
	\centering
	\includegraphics[width=\textwidth]{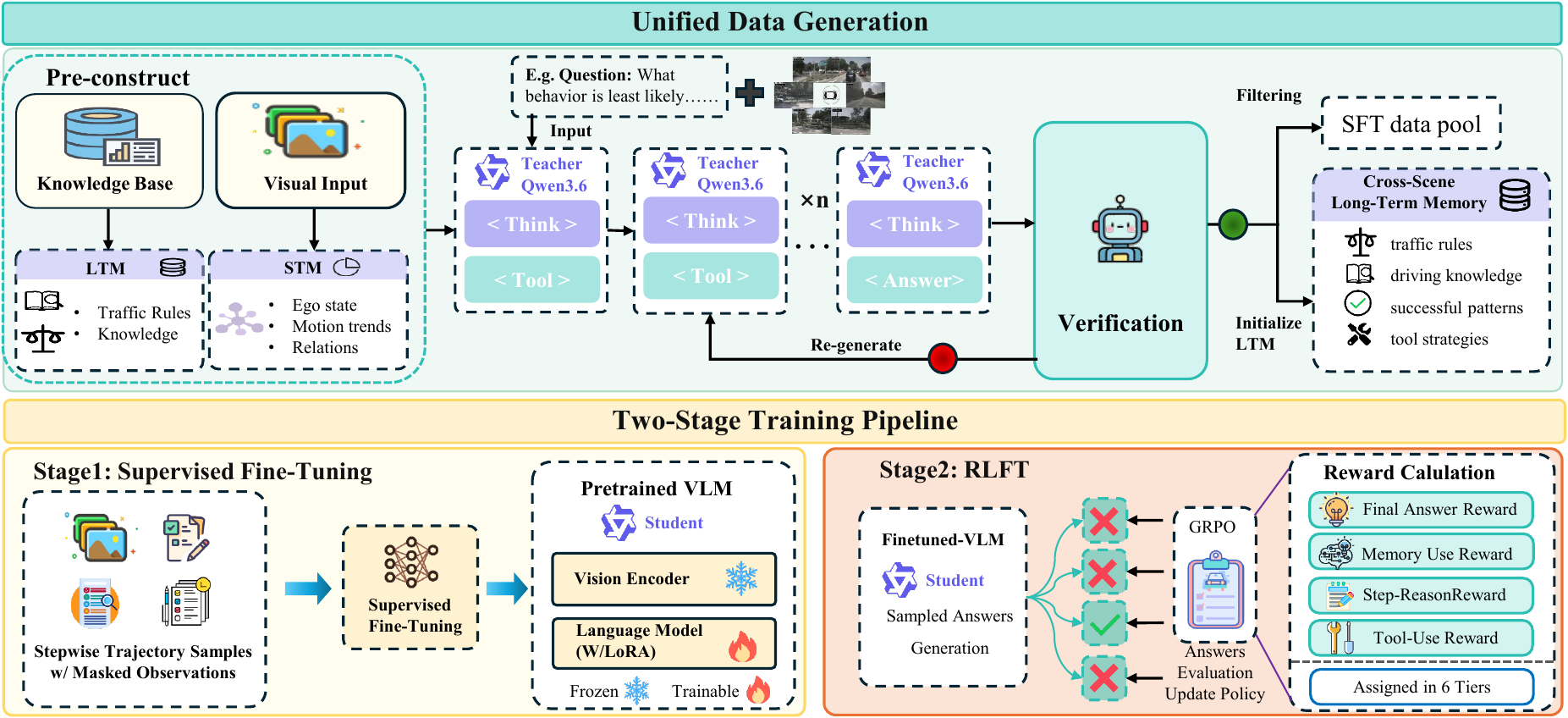}
	\caption{Data generation and two-stage post-training. A teacher VLM generates verified memory-tool trajectories via tool executors; stepwise SFT provides a tool-use cold start, followed by GRPO.}
	\label{fig:pipeline}
\end{figure*}
\subsubsection{Offline consolidation.}
The long-term memory remains read-only during online
reasoning. Let $\mathcal{C}$ denote a completed memory--tool
trajectory, formally defined in Sec.~\ref{sec:mtsr}, and let
$M_l^{(n)}$ denote the LTM before the $n$-th consolidation
round. We first extract reusable experience from trajectories
that pass verification and then consolidate the resulting
experience buffer:
\begin{equation}
	\small
	\mathcal{B}_n^{+}
	=
	\bigl\{
	\operatorname{Ext}(\mathcal{C})
	\mid
	\operatorname{Vrf}(\mathcal{C})=1
	\bigr\},
	\quad
	M_l^{(n+1)}
	=
	\Phi_{\mathrm{con}}
	\bigl(M_l^{(n)},\mathcal{B}_n^{+}\bigr).
	\label{eq:ltm_consolidation}
\end{equation}
$\operatorname{Ext}(\cdot)$ maps a completed trajectory to the
experience schema defined above, while
$\operatorname{Vrf}(\cdot)$ verifies final-answer correctness,
tool-call validity, and reasoning--observation consistency.
The resulting verified experiences are filtered, abstracted,
deduplicated, and merged by the offline consolidation operator
$\Phi_{\mathrm{con}}$.
For an existing entry $e_i$, we assign a utility credit
$g_i(\mathcal{C})\in\{0,1\}$ to each verified trajectory and
update its retrieval utility as
\begin{equation}
	w_i
	\leftarrow
	w_i+\eta\,g_i(\mathcal{C}),
	\label{eq:utility_update}
\end{equation}
where $\eta$ is the utility-update rate, and
$g_i(\mathcal{C})=1$ only when $e_i\in\mathcal{E}_K$ and the
verifier determines that it provided useful guidance to
$\mathcal{C}$; otherwise, $g_i(\mathcal{C})=0$. The complete offline consolidation procedure is summarized in Appendix~\ref{app:mem_ltm}.

\subsection{Memory-Tool Synergistic Reasoning}
\label{sec:mtsr}
Conditioned on $\mathcal{M}_s$ and $\mathcal{E}_K$, the model
evolves a transient scene belief through tool interaction.
Starting from
$x_0=(q,\mathcal{I},\mathcal{M}_s,\mathcal{E}_K)$, a complete
trajectory is
\begin{equation}
	\mathcal{C}=(x_0,s_1,\ldots,s_T,y), \qquad
	s_t=(h_t,a_t,o_t,u_t),
\end{equation}
where $h_t$, $a_t$, $o_t$, and $u_t$ denote the reasoning
state, action, tool observation, and revised scene belief,
respectively, with
$u_0=(\mathcal{M}_s,\mathcal{E}_K)$.

At step $t$, the policy generates the next reasoning state and
action from the interaction history:
\begin{equation}
	(h_t,a_t)\sim\pi_\theta(\cdot\mid x_0,\mathcal{H}_{<t}),
	\qquad
	a_t\in\mathcal{T}\cup\{\mathrm{finish}\},
\end{equation}
where $\mathcal{H}_{<t}=(s_1,\ldots,s_{t-1})$. Both memories
remain fixed within a query; only the transient scene belief
evolves as observations are incorporated. Once sufficient
evidence has been collected, the model emits
\texttt{finish} and generates
\begin{equation}
	y\sim\pi_\theta(\cdot\mid x_0,\mathcal{H}_{\leq T}).
\end{equation}

Thus, reasoning follows a
Think--Action--Observation--Update loop without a predefined
tool sequence: $\mathcal{M}_s$ describes the current scene,
$\mathcal{E}_K$ suggests relevant risks and strategies, and
each observation narrows the remaining uncertainty. Tool
interfaces are detailed in Appendix~A.3.

\subsection{Data Generation and Two-Stage Post-Training}
\label{sec:posttrain}
We construct verified memory-tool trajectories and train the
policy using stepwise SFT followed by trajectory-level GRPO, as shown in Fig.~\ref{fig:pipeline}.

\subsubsection{Multi-Step Teacher Rollout and Validation.}
To expose the student to the same memory-conditioned tool-use process, a teacher
policy $\pi_\phi$ starts from the initial context $x_0$, retrieves relevant
entries from an initialized long-term memory $M_l^{(0)}$, and generates
multi-step trajectories through the MTSR loop. Each trajectory is validated for
final-answer correctness, tool-call validity, and reasoning-observation
consistency. Verified ones form the training set $\mathcal{D}_{\mathrm{MTSR}}$,
which is also consolidated into $M_l^{(0)}$ (Algorithm~\ref{alg:consolidation})
to yield the student's initial long-term memory.

\subsubsection{Stepwise Supervised Fine-Tuning Warm-up.}
We perform stepwise SFT in two phases to progressively establish memory-tool
interaction. \textbf{Phase 1} removes executor-returned
observations and trains the model to select tools and generate arguments.
\textbf{Phase 2} continues from the Phase-1 checkpoint with complete memory and tool feedback to learn
observation-grounded reasoning.
Let $\mathcal{D}^{(p)}$ and $\mathcal{Y}^{(p)}(\mathcal{C})$
denote the training data and supervised token positions for
phase $p\in\{1,2\}$. The unified SFT objective is
\begin{equation}
	\mathcal{L}_{\mathrm{SFT}}^{(p)}(\theta)
	=
	-
	\mathbb{E}_{\mathcal{C}\sim\mathcal{D}^{(p)}}
	\left[
	\sum_{j\in\mathcal{Y}^{(p)}(\mathcal{C})}
	\log
	\pi_\theta
	\left(
	z_j^{(p)}
	\mid
	z_{<j}^{(p)}
	\right)
	\right].
	\label{eq:sft_objective}
\end{equation}
Here, $\mathcal{Y}^{(1)}$ contains tool-call tokens, whereas
$\mathcal{Y}^{(2)}$ contains model-generated reasoning,
actions, belief updates, and final answers. Executor-returned
observations are provided as context but excluded from the
training loss. The two-phase SFT warm-up prepares the model
for memory-conditioned reasoning and effective tool
integration prior to GRPO.

\subsubsection{Trajectory-Level GRPO.}

Starting from the Phase-2 checkpoint, GRPO samples $G$
trajectories per input. A correctness-gated reward jointly
evaluates final-answer correctness, relevant memory use, valid
tool interaction, and output format; auxiliary rewards activate
only for correct answers. We optimize
\begin{equation}
	\small
	\mathcal{L}_{\mathrm{GRPO}}(\theta)
	=
	-\mathbb{E}\Bigg[
	\frac{1}{G}
	\sum_{i=1}^{G}
	\frac{1}{|\mathcal{Y}(\mathcal{C}_i)|}
	\sum_{j\in\mathcal{Y}(\mathcal{C}_i)}
	\ell_{i,j}^{\mathrm{clip}}
	-
	\beta
	D_{\mathrm{KL}}
	\left(
	\pi_\theta
	\Vert
	\pi_{\mathrm{ref}}
	\right)
	\Bigg].
	\label{eq:grpo_objective}
\end{equation}
where $\ell_{i,j}^{\mathrm{clip}}$ uses the trajectory-level
group-relative advantage and $\pi_{\mathrm{ref}}$ is initialized
from the Phase-2 policy. Reward tiers and full optimization
details are provided in Appendix~B.3.

\section{Experiments}
\label{sec:exp}

\begin{figure*}[t]
	\centering
	\includegraphics[width=\textwidth]{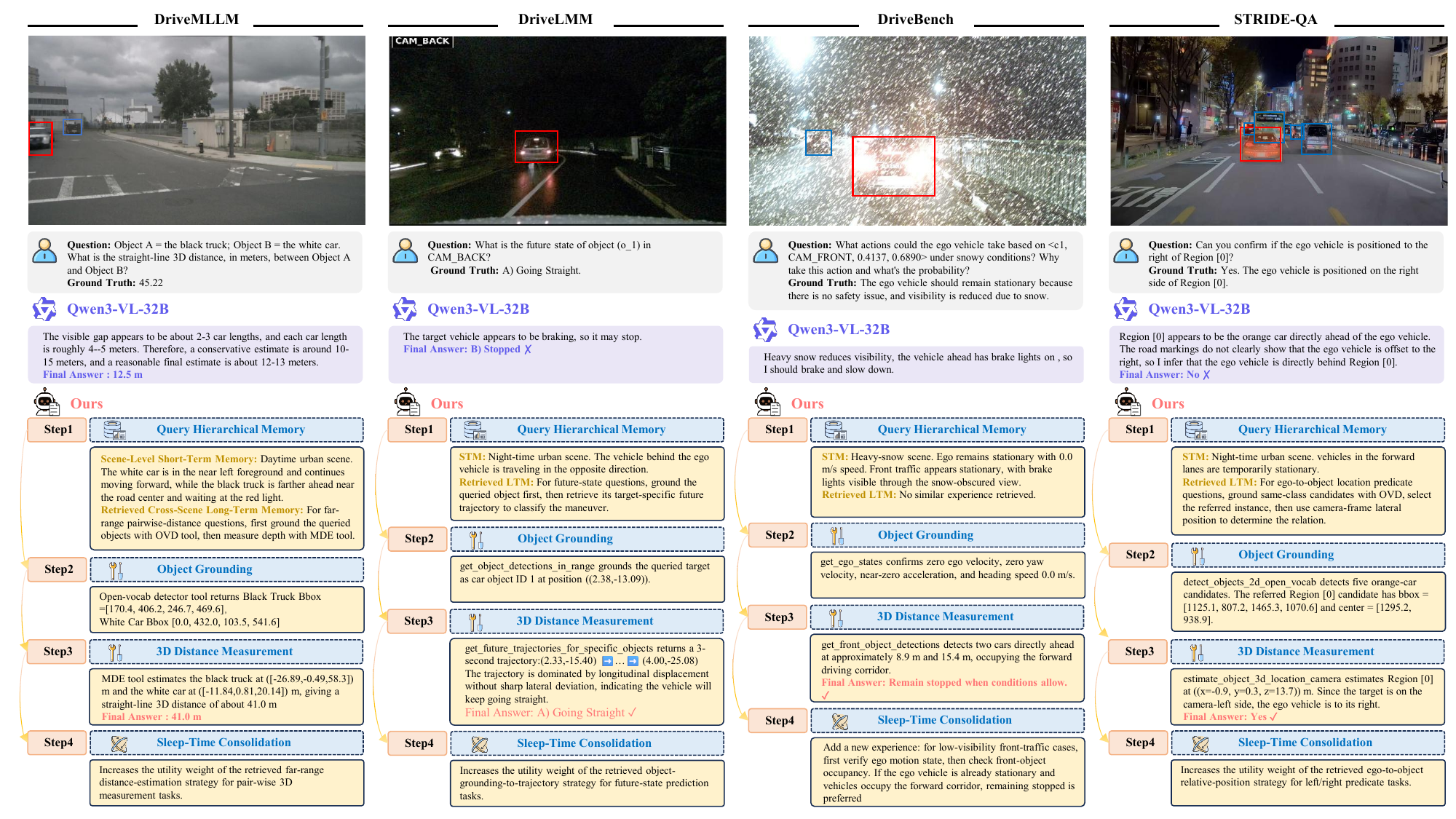}
	\caption{Qualitative Comparison with Qwen3-VL-32B on DriveMLLM, DriveLMM-o1, DriveBench and STRIDE-QA. All results are zero-shot except for DriveLMM.}
	\label{fig:qualitative_results}
\end{figure*}

We conduct experiments to answer the following four questions:

\textbf{Q1.} Can our framework improve both answer accuracy and reasoning quality over strong general-purpose and reasoning-enhanced VLM baselines? (\S\ref{sec:main_results})

\textbf{Q2.} How well does our framework generalize across benchmarks and challenge other tasks without further training? (\S\ref{sec:main_results})

\textbf{Q3.} How do hierarchical driving memory, two-stage post-training, and memory-guided proactive tool invocation contribute to model performance? (\S\ref{sec:ablation})

\textbf{Q4.} Can offline consolidation of LTM yield cumulative improvements without updating model parameters? (\S\ref{sec:analysis})

\begin{table*}[t]
\centering
\small
\setlength{\tabcolsep}{4pt}
\begin{tabular}{l ccc cc cc}
\toprule
\multirow{2}{*}{Vision-Language Models} & \multicolumn{3}{c}{Driving Metrics (\%) $\uparrow$} & \multicolumn{2}{c}{Scene Detail (\%) $\uparrow$} & \multicolumn{2}{c}{Overall (\%) $\uparrow$}\\
\cmidrule(lr){2-4}\cmidrule(lr){5-6}\cmidrule(lr){7-8}
& Risk Assess. & Rule Adh. & Scene Aware. & Relevance & Missing & Reason. & MCQ\\
\midrule
GPT-4o & 71.32 & 80.72 & 72.96 & 76.65 & 71.43 & 72.52 & 57.84\\
Ovis1.5-Gemma2-9B & 51.34 & 66.36 & 54.74 & 55.72 & 55.74 & 55.62 & 48.85\\
LLaVA-CoT & 57.62 & 69.01 & 60.84 & 62.72 & 60.67 & 61.41 & 49.27\\
InternVL2.5-8B & 69.02 & 78.43 & 71.52 & 75.80 & 70.54 & 71.62 & 54.87\\
Qwen2.5-VL-72B & 64.40 & 72.81 & 60.29 & 65.13 & 62.81 & 65.73 & 61.27\\
Qwen3-VL-8B & \underline{79.50} & 84.32 & 80.41 & \underline{84.43} & 75.32 & 77.76 & 55.54\\
Qwen2.5-VL-7B (base) & 46.44 & 60.45 & 51.02 & 50.15 & 52.19 & 51.77 & 37.81\\
DriveLMM-o1 & 73.01 & 81.56 & 75.39 & 79.42 & 74.49 & 75.24 & 62.36\\
AgentThink & \textbf{80.51} & \underline{84.98} & \underline{82.11} & \textbf{84.99} & \textbf{79.56} & \underline{79.68} & \underline{71.35}\\
\midrule
\textbf{Ours} & 76.95 & \textbf{86.28} & \textbf{84.05} & 77.38 & \underline{76.39} & \textbf{80.03} & \textbf{79.09}\\
\bottomrule
\end{tabular}
\caption{Performance comparison on the DriveLMM-o1 benchmark. \textbf{Bold} is best, \underline{underline} is second best.}
\label{tab:drivelmmo1}
\end{table*}

\subsection{Experimental Setup}
\label{sec:setup}

\subsubsection{Benchmarks and Evaluation Metrics.}
DriveLMM-o1~\cite{ishaqDriveLMMo1StepbystepReasoning2025} is the primary benchmark; we
report its overall reasoning score, five driving-specific
dimensions, and MCQ accuracy. DriveMLLM~\cite{guoDriveMLLMBenchmark2024}
evaluates zero-/one-shot generalization over eight fine-grained
tasks using their mean accuracy (AccS). STRIDE-QA
~\cite{ishiharaSTRIDEQA2026} reports localization success at 0--3\,s,
its mean (MLSR), and temporal localization consistency (TLC).
STSBench~\cite{fruhwirthSTSBench2025} evaluates spatio-temporal
understanding over 971 questions using category accuracy and a
question-weighted micro average. Full protocols are provided in
Appendix~C.

\subsubsection{Training, Inference and Implementation.}
We adapt Qwen2.5-VL-7B-Instruct~\cite{baiQwen25VL2025} using LoRA ($r=16$, $\alpha=32$) while freezing the
vision encoder. Qwen3.6-Plus generates 7,000 tool-executing
trajectories, of which 6,383 pass answer verification and
trajectory-quality filtering. Two-phase stepwise SFT first
learns valid tool selection and argument generation for
3 epochs, and then observation-grounded reasoning for
20 epochs, followed by trajectory-level GRPO which
samples $G=8$ responses per prompt. At inference, the model retrieves
the top-4 LTM entries and performs at most eight
Think-Action-Observation-Update turns with tool
execution. All experiments are conducted on 4×NVIDIA RTX PRO 6000 Blackwell GPUs (96GB memory), data construction and implementation details are given
in Appendix~\ref{app:data}.


\subsection{Main Results}
\label{sec:main_results}

\begin{table*}[t]
	\centering
	\setlength{\tabcolsep}{3pt}
	\resizebox{\textwidth}{!}{%
		\begin{tabular}{l ccccccccc | ccccccccc}
			\toprule
			\multirow{2}{*}{Model} & \multicolumn{9}{c}{Zero-shot} & \multicolumn{9}{c}{One-shot}\\
			\cmidrule(lr){2-10}\cmidrule(lr){11-19}
			& L/R & F/B & RHD & RD & PPos & BBox & CVD & CD & AccS
			& L/R & F/B & RHD & RD & PPos & BBox & CVD & CD & AccS\\
			\midrule
			GPT-4o
			& \textbf{91.72} & \underline{67.60} & 9.58 & 14.69 & 40.90 & 4.07 & 46.11 & \underline{70.65} & 43.16
			& \textbf{91.08} & \underline{69.37} & 36.51 & \underline{71.17} & 42.44 & 5.10 & 0.00 & \underline{63.88} & 47.44\\
			GPT-4o-mini
			& 67.67 & 50.13 & \underline{70.44} & 0.00 & 29.28 & 3.78 & 0.00 & 46.40 & 33.46
			& 66.00 & 48.95 & \textbf{83.02} & 58.47 & 25.71 & 3.97 & 52.73 & 55.23 & 49.26\\
			LLaVA-ov-72B
			& 85.42 & 49.48 & 13.76 & 45.27 & 16.46 & 0.00 & 42.97 & 27.09 & 35.06
			& 79.12 & 62.97 & 49.26 & 68.04 & 28.57 & 2.20 & \underline{53.12} & 60.90 & 50.52\\
			Qwen2.5-VL-7B
			& 76.55 & 55.24 & 7.14 & 17.11 & 55.97 & 38.31 & 55.94 & 51.52 & 44.72
			& 80.30 & 53.14 & 36.96 & 39.13 & 62.69 & 22.63 & 49.88 & 48.32 & 49.13\\
			Qwen + CoT
			& 87.06 & 63.09 & 16.69 & 22.56 & 52.51 & 38.87 & \underline{76.90} & 38.71 & 49.55
			& 86.35 & 59.95 & 43.29 & 31.81 & 53.64 & 26.93 & 51.02 & 42.30 & 49.41\\
			Qwen + DirectTool
			& 78.95 & 48.96 & 58.43 & \underline{67.57} & 58.20 & 42.22 & 51.76 & 51.38 & \underline{57.18}
			& 84.57 & 55.50 & 67.32 & 59.54 & \textbf{85.58} & 26.07 & 52.34 & 53.25 & 60.52\\
			AgentThink
			& 82.33 & 54.40 & 56.14 & 61.45 & \textbf{70.45} & \underline{56.23} & 23.09 & 51.60 & 56.96
			& 78.71 & 48.46 & 60.64 & 60.71 & 72.36 & \underline{64.46} & 52.26 & 52.04 & \underline{61.21}\\
			\textbf{Ours}
			& \underline{89.28} & \textbf{78.02} & \textbf{73.94} & \textbf{70.50} & \underline{67.15} & \textbf{63.27} & \textbf{77.47} & \textbf{75.26} & \textbf{74.36}
			& \underline{87.16} & \textbf{83.12} & \underline{80.89} & \textbf{74.29} & \underline{76.27} & \textbf{67.92} & \textbf{87.86} & \textbf{87.19} & \textbf{80.59}\\
			\bottomrule
	\end{tabular}}
	\caption{Zero-shot and one-shot results on DriveMLLM (\%). \textbf{Bold} best, \underline{underline} second best, within each setting.}
	\label{tab:drivemllm}
\end{table*}

\subsubsection{Results on DriveLMM-o1.}
%
Table~\ref{tab:drivelmmo1} answers Q1. Our framework achieves
the best reasoning score of 80.03 and MCQ accuracy of
79.09\%, improving the Qwen2.5-VL-7B backbone by 28.26
and 41.28 points, respectively. It also exceeds the same-scale
AgentThink~\cite{qianAgentThinkUnifiedFramework2025} by 7.74 MCQ points and ranks first on Rule
Adherence and Scene Awareness.
DriveAgent-R1~\cite{zhengDriveAgentR12025} is the closest concurrent method in
active perception. However, its checkpoints
and evaluation assets have not been fully released, preventing a
faithful comparison under the same protocol. Among reproducible
tool-augmented baselines, AgentThink injects pre-computed tool outputs
into single-pass generation, whereas our framework selects and
executes tools online within a memory-conditioned interaction
loop. This comparison demonstrates a favorable
accuracy-reasoning trade-off, while the effects of components are examined in \S\ref{sec:ablation}. 

\subsubsection{Cross-Benchmark Generalization.}

\begin{table}[t]
	\centering
	\small
	\setlength{\tabcolsep}{3.5pt}
	\resizebox{\columnwidth}{!}{
		\begin{tabular}{lcccccc}
			\toprule
			Model
			& LSR$_{0\mathrm{s}}$
			& LSR$_{1\mathrm{s}}$
			& LSR$_{2\mathrm{s}}$
			& LSR$_{3\mathrm{s}}$
			& MLSR
			& TLC \\
			\midrule
			GPT-4o
			& 18.1 & 6.6 & 6.1 & 7.6 & 9.6 & 0.7 \\
			\midrule
			InternVL2.5-8B
			& \textbf{2.4} & 1.0 & 1.7 & 0.7 & 1.5 & 0.0 \\
			
			Qwen3-VL-8B
			& 1.0 & 3.2 & 4.4 & 1.0 & 2.4 & 0.0 \\
			
			SpatialRGPT-VILA-1.5-8B
			& 0.5 & 0.2 & 0.2 & 0.0 & 0.2 & 0.0 \\
			
			Cosmos-Reason1-7B
			& 1.5 & 3.2 & 2.0 & 1.5 & 2.0 & 0.0 \\
			
			\textbf{Ours}
			& 2.0 & \textbf{3.4} & \textbf{5.6}
			& \textbf{4.4} & \textbf{3.9} & \textbf{0.2} \\
			\bottomrule
		\end{tabular}
	}
	\caption{
		Zero-shot cross-dataset results on STRIDE-QA. GPT-4o is included as a proprietary reference; bold denotes the best result
		among the remaining models.
	}
	\label{tab:strideqa}
\end{table}

Turning to Q2, we evaluate all models without additional
training or memory updates. On DriveMLLM,
Table~\ref{tab:drivemllm} shows that our framework obtains zero-/one-shot AccS scores of 74.36/80.59 on
DriveMLLM, versus 56.96/61.21 for AgentThink and
57.18/60.52 for direct tool injection
(Table~\ref{tab:drivemllm}). Gains concentrate on grounding
and metric-geometry tasks. On unseen STRIDE-QA, it achieves
the best open-model MLSR of 3.9 and the strongest
longer-horizon results, reaching 5.6/4.4 LSR at 2/3\,s
(Table~\ref{tab:strideqa}). These results support the transfer
of STM scene states and LTM tool-use experience.

\subsubsection{Qualitative Analysis.}
%
We additionally test our framework on DriveBench~\cite{xieAreVLMsReady2025a} solely to examine robustness under visual corruption. As shown in Figure~\ref{fig:qualitative_results}, on
long-range distance estimation, future-state prediction, adverse-weather
decision making, and ego-object spatial reasoning, including degraded
low-visibility scenes, our model actively grounds its answer in tool evidence
and reuses retrieved experience, rather than guessing from appearance as
Qwen3-VL-32B does (e.g., $41.0$\,m vs.\ $12.5$\,m against a $45.22$\,m ground
truth).

\begin{figure}[t] 
	\centering
	\begin{subfigure}[b]{0.44\columnwidth}
		\centering
		\includegraphics[width=\textwidth]{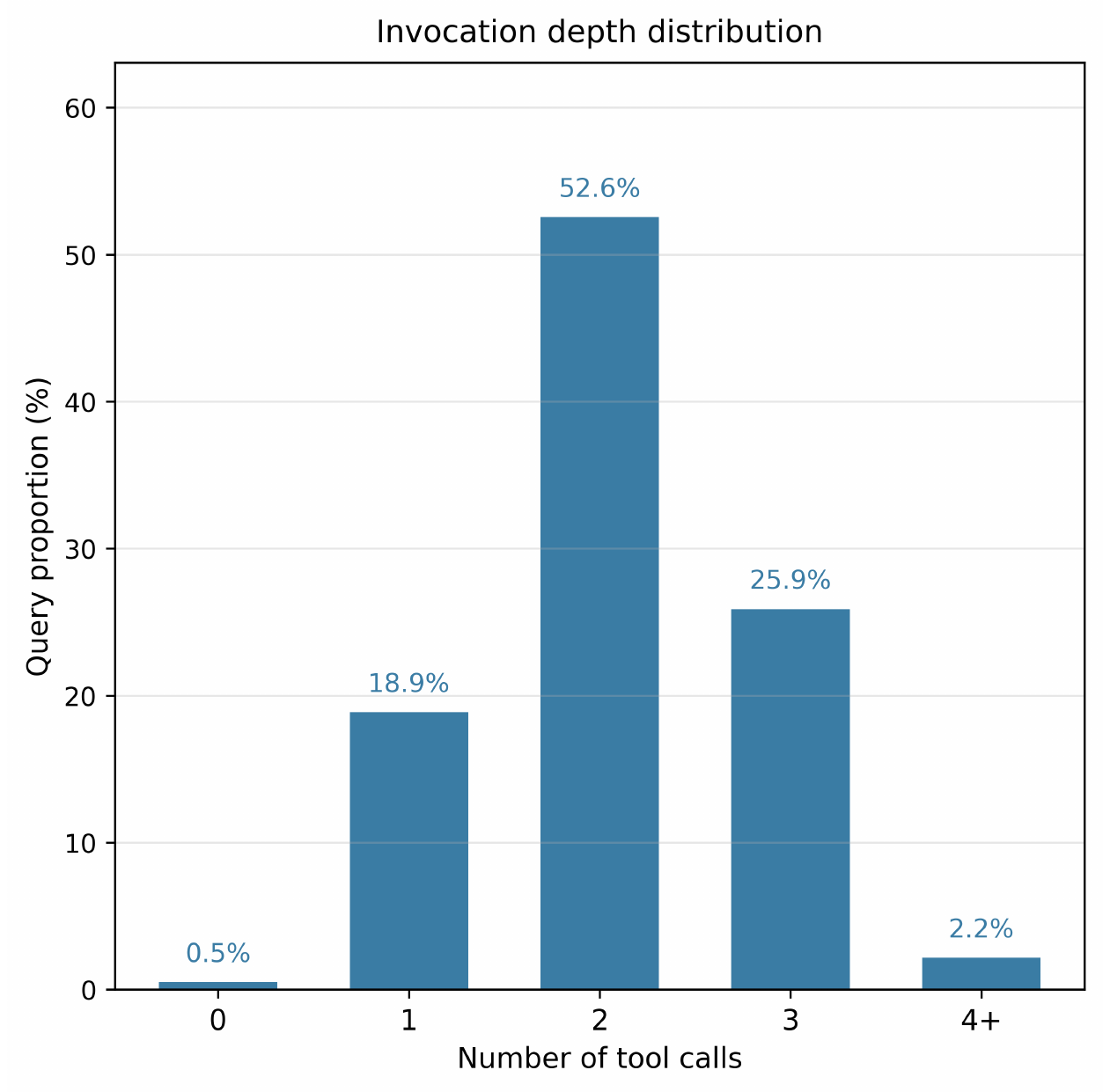}
		\caption{Tool-call depth.}
		\label{fig:invocation_depth}
	\end{subfigure}
	\hfill
	\begin{subfigure}[b]{0.55\columnwidth}
		\centering
		\includegraphics[width=\textwidth]{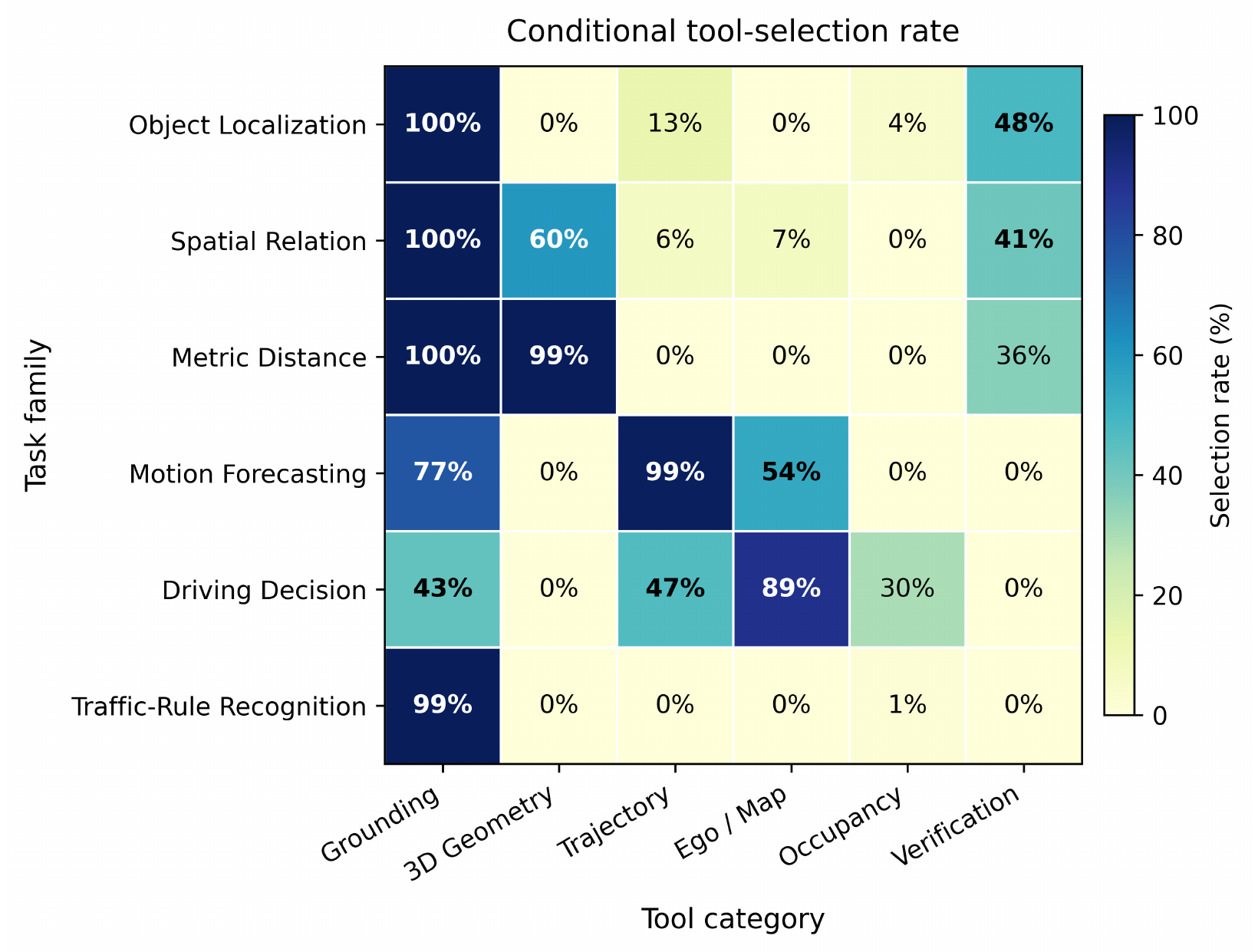}
		\caption{Conditional tool selection.}
		\label{fig:tool_alignment}
	\end{subfigure}
	
	\vspace{4pt}
	\begin{subfigure}[b]{0.5\columnwidth}
		\centering
		\includegraphics[width=\textwidth]{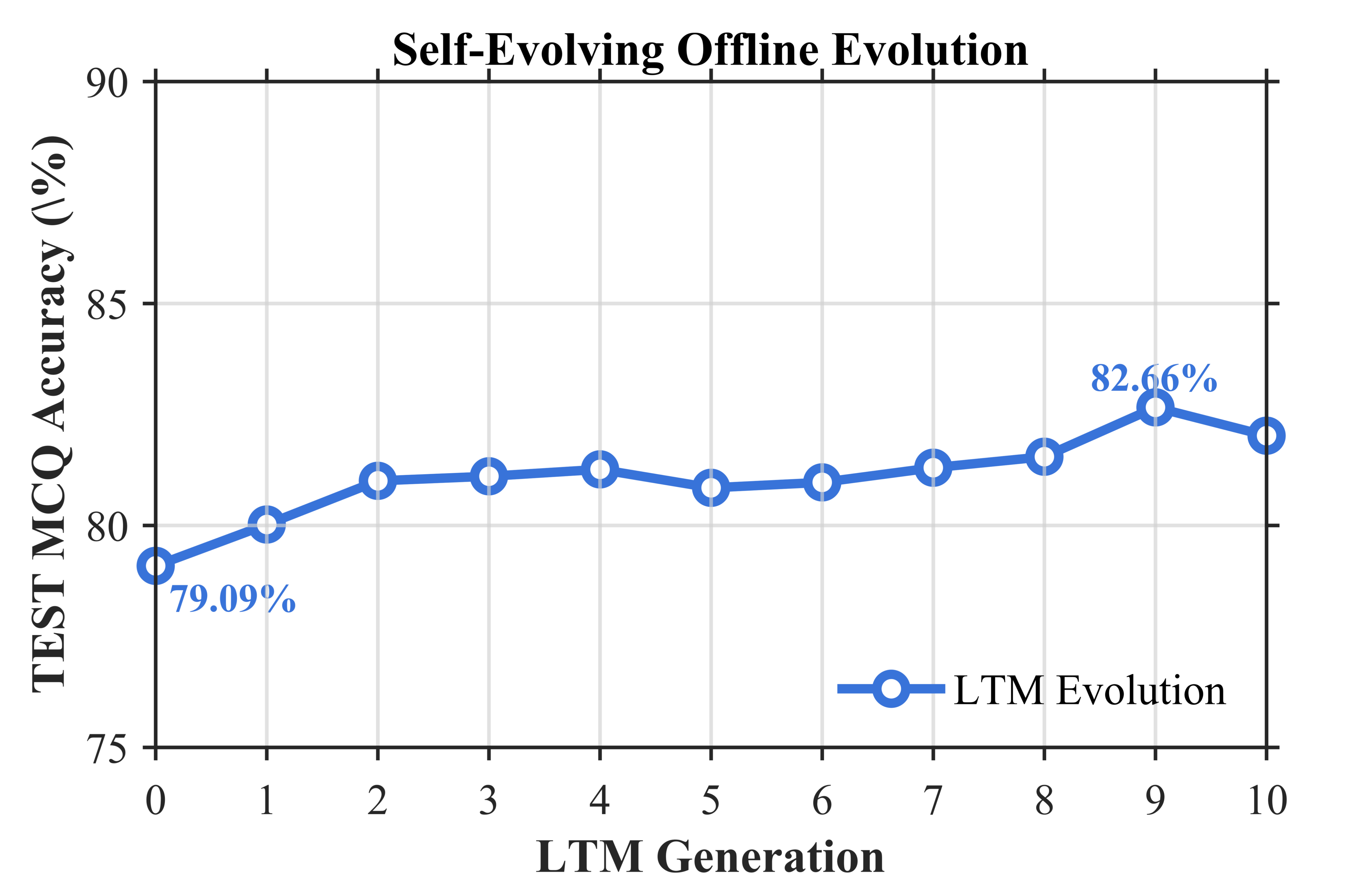}
		\caption{Long-term memory self-evolution.}
		\label{fig:self_evolution}
	\end{subfigure}
	
	\caption{Behavioral and evolutionary analysis. (a) Distribution of
		interaction depth. (b) Task-conditioned tool selection. (c) Test MCQ over
		sleep-time LTM consolidation rounds (frozen model).}
	\label{fig:tool_behavior}
\end{figure}

\subsection{Ablation Study}
\label{sec:ablation}

\begin{table}[!t]
\centering
\setlength{\tabcolsep}{3pt}
\resizebox{\columnwidth}{!}{%
\begin{tabular}{l ccccc}
	\toprule
	Method
	& Overall
	& Ego
	& Ego-Ag.
	& Ag.
	& Ag.-Ag. \\
	\midrule
	GPT-4o
	& \underline{50.25}
	& 63.63
	& \textbf{75.75}
	& 45.59
	& 43.38 \\
	
	InternVL3-8B
	& 42.85
	& 34.36
	& \underline{51.41}
	& 46.69
	& 39.52 \\
	\midrule
    Qwen2.5-VL 7B
	& 31.38
	& 26.47
	& 34.17
	& 32.74
	& 30.73 \\
	
    \;\;+ 6 frames input
	& 35.64
	& 35.62
	& 26.64
	& 36.58
	& 37.47 \\

	\;\;+ static STM
	& \underline{50.36}
	& \underline{81.37}
	& 37.50
	& \underline{52.17}
	& \underline{45.16} \\
	
	\;\;+ STM
	& \textbf{55.61}
	& \textbf{84.31}
	& 45.00
	& \textbf{61.23}
	& \textbf{47.41} \\
	\bottomrule
\end{tabular}
}
\caption{Zero-shot effect of the short-term memory (STM) on STSBench temporal-dynamic perception (\%). \textbf{Bold} best, \underline{underline} second best. Rel.\ = Relation, Ag.\ = Agent.}
\label{tab:stsbench}
\end{table}

Turning to \textbf{Q3}, we examine how post-training, hierarchical memory,
and structured temporal context contribute to the final performance. 

\subsubsection{Post-training stages.}
Turning to Q3, Table~\ref{tab:ablations} shows two-stage SFT raises
reasoning/MCQ from $51.77/37.81$ to $73.91/67.06$, and adding GRPO yields our
full model at $80.03/79.09$. GRPO on the base alone reaches only
$73.85/64.49$, even below SFT in MCQ, showing the SFT cold start is a
necessary prerequisite. SFT establishes valid interaction,
while GRPO refines complete trajectories once that foundation is in place.

\subsubsection{Hierarchical memory.} We isolate the two memory levels while
retaining the trained model and tool executor. As shown in Table~\ref{tab:ablations},
the no-memory variant reaches $72.16\%$ MCQ; adding LTM alone improves it to
$75.20\%$, STM alone to $76.91\%$, and combining both yields the best $79.09\%$.
STM contributes more, as most DriveLMM-o1 questions depend on the current scene
state, while LTM adds a smaller but consistent gain from reusable experience;
their combination confirms the complementary roles of the two memories.

\subsubsection{Temporal Representation Ablation.}
Table~\ref{tab:stsbench} compares temporal representations on
STSBench. Raw six-frame input provides a modest gain over the
base model (35.64\% vs.\ 31.38\%), whereas static and
temporal STM reach 50.36\% and 55.61\%, respectively.
Explicit temporal-state organization is therefore more
effective than increasing visual context.


\begin{table}[t]
	\centering
	\small
	\setlength{\tabcolsep}{4pt}
	\begin{tabular}{@{}c@{\hspace{1.4em}}c@{}}
		\begin{tabular}[t]{@{}lcc@{}}
			\toprule
			Variant & Reason. & MCQ\\
			\midrule
			Base            & 51.77 & 37.81\\
			\;+ SFT          & 73.91 & 67.06\\
			\;+ GRPO		 & 73.85 & 64.49\\
			\;Ours  & \textbf{80.03} & \textbf{79.09}\\
			\bottomrule
		\end{tabular}
		&
		\begin{tabular}[t]{@{}lc@{}}
			\toprule
			Memory & MCQ\\
			\midrule
			w/o memory      & 72.16\\
			\;+ LTM          & 75.20\\
			\;+ STM          & 76.91\\
			\;+ STM + LTM    & \textbf{79.09}\\
			\bottomrule
		\end{tabular}
		\\[3pt]
		{\small (a) Post-training stages.} & {\small (b) Memory components.}\\
	\end{tabular}
	\caption{Ablations on DriveLMM-o1 (\%, $\uparrow$). \textbf{Bold} best. Memory components are added on the trained, tool-equipped no-memory model.}
	\label{tab:ablations}
\end{table}

\subsection{Behavioral and Evolutionary Analysis}
\label{sec:analysis}

\subsubsection{Online tool behavior.}
Figures~\ref{fig:invocation_depth} and~\ref{fig:tool_alignment} characterize online evidence
acquisition using benchmark-balanced macro averages. Overall,
97.4\% of queries use one to three tools, with an average of
2.10 calls, while no-tool answering occurs in only 0.5\% of
cases. Within the admissible tool sets, metric-distance,
motion-forecasting, and driving-decision questions
predominantly invoke 3D geometry, trajectory, and ego/map
tools, respectively. Tool execution succeeds in 99.74\% of
calls, with only 0.45\% exact duplicates. Moreover, 98.94\%
of trajectories terminate normally; 0.65\% stop without a
further action, 0.39\% terminate after a tool error, and only
0.02\% reach the maximum-turn limit. These results show
task-dependent, variable-length, and reliable interaction
rather than a fixed invocation schedule.

\subsubsection{Offline LTM evolution.}
To answer Q4, we consolidate successful trajectories
while freezing all model parameters. Only training-split and
additional non-benchmark scenes are used; no test information enters. Figure~\ref{fig:self_evolution} shows MCQ
increasing from 79.09\% to 82.66\% by generation~9, a
3.57-point gain without parameter updates. The slight
generation-10 decline indicates saturation, demonstrating
non-parametric improvement through accumulated experience.

\section{Conclusion}
\label{sec:conclusion}

We presented a hierarchical memory-tool framework that
combines foresight with hindsight, and adaptive evidence acquisition for autonomous driving reasoning. The experimental results highlight the importance of moving autonomous
driving VLMs toward adaptive
systems that can accumulate experience and interact with external
evidence. Future work will explore more scalable consolidation and extension toward open-world closed-loop
driving.

\clearpage
\bibliography{references}

\begin{thebibliography}{29}
\providecommand{\natexlab}[1]{#1}

\bibitem[{Bai et~al.(2025)Bai, Chen, Liu, Wang, Ge, Song, Dang, Wang, Wang,
  Tang et~al.}]{baiQwen25VL2025}
Bai, S.; Chen, K.; Liu, X.; Wang, J.; Ge, W.; Song, S.; Dang, K.; Wang, P.;
  Wang, S.; Tang, J.; et~al. 2025.
\newblock {Qwen2.5-VL} Technical Report.
\newblock arXiv:2502.13923.

\bibitem[{Fruhwirth-Reisinger et~al.(2025)Fruhwirth-Reisinger, Mali{\'c}, Lin,
  Schinagl, Schulter, and Possegger}]{fruhwirthSTSBench2025}
Fruhwirth-Reisinger, C.; Mali{\'c}, D.; Lin, W.; Schinagl, D.; Schulter, S.;
  and Possegger, H. 2025.
\newblock {STSBench}: A Spatio-Temporal Scenario Benchmark for Multi-Modal
  Large Language Models in Autonomous Driving.
\newblock In \emph{Advances in Neural Information Processing Systems 38
  (NeurIPS), Datasets and Benchmarks Track}.

\bibitem[{Fu et~al.(2025)Fu, Zhang, Zhao, Cui, Liang, Zhang, Zhang, Xie, Wang,
  and Bai}]{fuORIONHolisticEndtoend2025}
Fu, H.; Zhang, D.; Zhao, Z.; Cui, J.; Liang, D.; Zhang, C.; Zhang, D.; Xie, H.;
  Wang, B.; and Bai, X. 2025.
\newblock {ORION}: A Holistic End-to-End Autonomous Driving Framework by
  Vision-Language Instructed Action Generation.
\newblock In \emph{Proceedings of the IEEE/CVF International Conference on
  Computer Vision (ICCV)}.

\bibitem[{Gao et~al.(2025)Gao, Piccinini, Brusnicki, Zhang, and
  Betz}]{gaoNuRiskVisualQuestion2025}
Gao, Y.; Piccinini, M.; Brusnicki, R.; Zhang, Y.; and Betz, J. 2025.
\newblock {NuRisk}: A Visual Question Answering Dataset for Agent-Level Risk
  Assessment in Autonomous Driving.
\newblock arXiv:2509.25944.

\bibitem[{Gholami et~al.(2025)Gholami, Rezaei, Weimin, Mao, Zhou, Zhang, and
  Akbari}]{gholamiSpatialReasoning2025}
Gholami, M.; Rezaei, A.; Weimin, Z.; Mao, S.; Zhou, S.; Zhang, Y.; and Akbari,
  M. 2025.
\newblock Spatial Reasoning with Vision-Language Models in Ego-Centric
  Multi-View Scenes.
\newblock arXiv:2509.06266.

\bibitem[{Guo et~al.(2024)Guo, Zhang, Duan, He, Zhang, Liu, and
  Chen}]{guoDriveMLLMBenchmark2024}
Guo, X.; Zhang, R.; Duan, Y.; He, Y.; Zhang, C.; Liu, S.; and Chen, L. 2024.
\newblock {DriveMLLM}: A Benchmark for Spatial Understanding with Multimodal
  Large Language Models in Autonomous Driving.
\newblock arXiv:2411.13112.

\bibitem[{Hwang et~al.(2025)Hwang, Xu, Lin, Hung, Ji, Choi, Huang, He,
  Covington, Sapp, Zhou, Guo, Anguelov, and Tan}]{hwangEMMA2024}
Hwang, J.-J.; Xu, R.; Lin, H.; Hung, W.-C.; Ji, J.; Choi, K.; Huang, D.; He,
  T.; Covington, P.; Sapp, B.; Zhou, Y.; Guo, J.; Anguelov, D.; and Tan, M.
  2025.
\newblock {EMMA}: End-to-End Multimodal Model for Autonomous Driving.
\newblock \emph{Transactions on Machine Learning Research}.

\bibitem[{Ishaq et~al.(2025)Ishaq, Lahoud, More, Thawakar, Thawkar,
  Dissanayake, Ahsan, Li, Khan, Cholakkal, Laptev, Anwer, and
  Khan}]{ishaqDriveLMMo1StepbystepReasoning2025}
Ishaq, A.; Lahoud, J.; More, K.; Thawakar, O.; Thawkar, R.; Dissanayake, D.;
  Ahsan, N.; Li, Y.; Khan, F.~S.; Cholakkal, H.; Laptev, I.; Anwer, R.~M.; and
  Khan, S. 2025.
\newblock {DriveLMM-o1}: A Step-by-Step Reasoning Dataset and Large Multimodal
  Model for Driving Scenario Understanding.
\newblock arXiv:2503.10621.

\bibitem[{Ishihara et~al.(2025)Ishihara, Sasaki, Takahashi, Shiono, and
  Yamaguchi}]{ishiharaSTRIDEQA2026}
Ishihara, K.; Sasaki, K.; Takahashi, T.; Shiono, D.; and Yamaguchi, Y. 2025.
\newblock {STRIDE-QA}: Visual Question Answering Dataset for Spatiotemporal
  Reasoning in Urban Driving Scenes.
\newblock arXiv:2508.10427.

\bibitem[{Jiang et~al.(2024)Jiang, Chen, Liao, Zhang, Yin, Zhang, Huang, Liu,
  and Wang}]{jiangSennaBridgingLarge2024a}
Jiang, B.; Chen, S.; Liao, B.; Zhang, X.; Yin, W.; Zhang, Q.; Huang, C.; Liu,
  W.; and Wang, X. 2024.
\newblock {Senna}: Bridging Large Vision-Language Models and End-to-End
  Autonomous Driving.
\newblock arXiv:2410.22313.

\bibitem[{Liu et~al.(2026)Liu, Ye, Zhang, and Qi}]{liuCritiqueDriveVLM2026}
Liu, Z.; Ye, H.; Zhang, X.; and Qi, M. 2026.
\newblock {CritiqueDriveVLM}: From Verifier-Guided Reinforcement Learning to
  Latent Thought Distillation for Autonomous Driving.
\newblock arXiv:2607.04179.

\bibitem[{Mao et~al.(2024)Mao, Ye, Qian, Pavone, and
  Wang}]{maoLanguageAgentAutonomous2024}
Mao, J.; Ye, J.; Qian, Y.; Pavone, M.; and Wang, Y. 2024.
\newblock A Language Agent for Autonomous Driving.
\newblock arXiv:2311.10813.

\bibitem[{Marcu et~al.(2024)Marcu, Chen, H{\"u}nermann, Karnsund, Hanotte,
  Chidananda, Nair, Badrinarayanan, Kendall, Shotton, Arani, and
  Sinavski}]{marcuLingoQA2024}
Marcu, A.-M.; Chen, L.; H{\"u}nermann, J.; Karnsund, A.; Hanotte, B.;
  Chidananda, P.; Nair, S.; Badrinarayanan, V.; Kendall, A.; Shotton, J.;
  Arani, E.; and Sinavski, O. 2024.
\newblock {LingoQA}: Visual Question Answering for Autonomous Driving.
\newblock In \emph{Proceedings of the European Conference on Computer Vision
  (ECCV)}. Cham: Springer.

\bibitem[{Qian et~al.(2025)Qian, Jiang, Zhong, Luo, Huang, Zhu, Jiang, Yang,
  Fu, Miao, Shi, Lim, Liu, Zhou, Yu, Hu, Li, Chen, Ye, Sun, and
  Yang}]{qianAgentThinkUnifiedFramework2025}
Qian, K.; Jiang, S.; Zhong, Y.; Luo, Z.; Huang, Z.; Zhu, T.; Jiang, K.; Yang,
  M.; Fu, Z.; Miao, J.; Shi, Y.; Lim, H.~Z.; Liu, L.; Zhou, T.; Yu, H.; Hu, Y.;
  Li, G.; Chen, G.; Ye, H.; Sun, L.; and Yang, D. 2025.
\newblock {AgentThink}: A Unified Framework for Tool-Augmented Chain-of-Thought
  Reasoning in Vision-Language Models for Autonomous Driving.
\newblock arXiv:2505.15298.

\bibitem[{Qian et~al.(2024)Qian, Chen, Zhuo, Jiao, and
  Jiang}]{qianNuScenesQA2024}
Qian, T.; Chen, J.; Zhuo, L.; Jiao, Y.; and Jiang, Y.-G. 2024.
\newblock {NuScenes-QA}: A Multi-Modal Visual Question Answering Benchmark for
  Autonomous Driving Scenario.
\newblock In \emph{Proceedings of the Thirty-Eighth {AAAI} Conference on
  Artificial Intelligence}, 4542--4550. Washington, DC: {AAAI} Press.

\bibitem[{Sima et~al.(2024)Sima, Renz, Chitta, Chen, Zhang, Xie,
  Bei{\ss}wenger, Luo, Geiger, and Li}]{simaDriveLM2024}
Sima, C.; Renz, K.; Chitta, K.; Chen, L.; Zhang, H.; Xie, C.; Bei{\ss}wenger,
  J.; Luo, P.; Geiger, A.; and Li, H. 2024.
\newblock {DriveLM}: Driving with Graph Visual Question Answering.
\newblock In \emph{Proceedings of the European Conference on Computer Vision
  (ECCV)}. Cham: Springer.

\bibitem[{Tao et~al.(2026)Tao, Taghavi, Filev, Langari, and
  Pandey}]{taoNaviDriveVLMDecouplingHighlevel2026}
Tao, X.; Taghavi, P.; Filev, D.; Langari, R.; and Pandey, G. 2026.
\newblock {NaviDriveVLM}: Decoupling High-Level Reasoning and Motion Planning
  for Autonomous Driving.
\newblock arXiv:2603.07901.

\bibitem[{Tian et~al.(2024)Tian, Gu, Li, Liu, Wang, Zhao, Zhan, Jia, Lang, and
  Zhao}]{tianDriveVLM2024}
Tian, X.; Gu, J.; Li, B.; Liu, Y.; Wang, Y.; Zhao, Z.; Zhan, K.; Jia, P.; Lang,
  X.; and Zhao, H. 2024.
\newblock {DriveVLM}: The Convergence of Autonomous Driving and Large
  Vision-Language Models.
\newblock In \emph{Proceedings of the 8th Conference on Robot Learning (CoRL)}.

\bibitem[{Wang et~al.(2025{\natexlab{a}})Wang, Liu, Jiang, Wang, Jiao, Chu,
  Gao, and Chen}]{wangRADRetrievalaugmentedDecisionmaking2025}
Wang, Y.; Liu, Q.; Jiang, Z.; Wang, T.; Jiao, J.; Chu, H.; Gao, B.; and Chen,
  H. 2025{\natexlab{a}}.
\newblock {RAD}: Retrieval-Augmented Decision-Making of Meta-Actions with
  Vision-Language Models in Autonomous Driving.
\newblock In \emph{Proceedings of the IEEE/CVF Conference on Computer Vision
  and Pattern Recognition (CVPR) Workshops}.

\bibitem[{Wang et~al.(2025{\natexlab{b}})Wang, Luo, Bai, Cao, Che, Chen, Chen,
  Diamond, Ding, Ding et~al.}]{nvidiaAlpamayoR12025}
Wang, Y.; Luo, W.; Bai, J.; Cao, Y.; Che, T.; Chen, K.; Chen, Y.; Diamond, J.;
  Ding, Y.; Ding, W.; et~al. 2025{\natexlab{b}}.
\newblock {Alpamayo-R1}: Bridging Reasoning and Action Prediction for
  Generalizable Autonomous Driving in the Long Tail.
\newblock arXiv:2511.00088.

\bibitem[{Xie et~al.(2025)Xie, Kong, Dong, Sima, Zhang, Chen, Liu, and
  Pan}]{xieAreVLMsReady2025a}
Xie, S.; Kong, L.; Dong, Y.; Sima, C.; Zhang, W.; Chen, Q.~A.; Liu, Z.; and
  Pan, L. 2025.
\newblock Are {VLMs} Ready for Autonomous Driving? An Empirical Study from the
  Reliability, Data, and Metric Perspectives.
\newblock arXiv:2501.04003.

\bibitem[{Yang et~al.(2026)Yang, Huang, Huang, Liu, and
  Yang}]{yangJudgeThenDrive2026}
Yang, L.; Huang, J.; Huang, Z.; Liu, S.; and Yang, H. 2026.
\newblock Judge, Then Drive: A Critic-Centric Vision Language Action Framework
  for Autonomous Driving.
\newblock arXiv:2604.27366.

\bibitem[{Ye et~al.(2025)Ye, Qi, Liu, Liu, and
  Ma}]{yeSafeDriveRAGSafeAutonomous2025}
Ye, H.; Qi, M.; Liu, Z.; Liu, L.; and Ma, H. 2025.
\newblock {SafeDriveRAG}: Towards Safe Autonomous Driving with Knowledge
  Graph-Based Retrieval-Augmented Generation.
\newblock arXiv:2507.21585.

\bibitem[{Yu et~al.(2025)Yu, Lee, Kim, Shin, Park, Ryu, Jung, and
  Shim}]{yuWaymoQA2025}
Yu, S.; Lee, S.; Kim, N.; Shin, J.; Park, J.; Ryu, W.; Jung, R.; and Shim, H.
  2025.
\newblock {WaymoQA}: A Multi-View Visual Question Answering Dataset for
  Safety-Critical Reasoning in Autonomous Driving.
\newblock arXiv:2511.20022.

\bibitem[{Yuan et~al.(2024)Yuan, Sun, Omeiza, Zhao, Newman, Kunze, and
  Gadd}]{yuanRAGdriverGeneralisableDriving2026}
Yuan, J.; Sun, S.; Omeiza, D.; Zhao, B.; Newman, P.; Kunze, L.; and Gadd, M.
  2024.
\newblock {RAG-Driver}: Generalisable Driving Explanations with
  Retrieval-Augmented In-Context Learning in Multi-Modal Large Language Model.
\newblock In \emph{Proceedings of Robotics: Science and Systems (RSS)}.

\bibitem[{Zeng et~al.(2025)Zeng, Chang, Xie, Liu, Bai, Pan, Xu, Wei, and
  Guo}]{zengFutureSightDriveThinkingVisually2025}
Zeng, S.; Chang, X.; Xie, M.; Liu, X.; Bai, Y.; Pan, Z.; Xu, M.; Wei, X.; and
  Guo, N. 2025.
\newblock {FutureSightDrive}: Thinking Visually with Spatio-Temporal {CoT} for
  Autonomous Driving.
\newblock In \emph{Advances in Neural Information Processing Systems 38
  (NeurIPS)}.
\newblock Spotlight.

\bibitem[{Zhang et~al.(2026)Zhang, Zheng, Wang, Xu, Deng, Chen, Chen, Zhang,
  and Huang}]{zhangOmniDriveR1ReinforcementdrivenInterleaved2026}
Zhang, Z.; Zheng, H.; Wang, Y.; Xu, L.; Deng, T.; Chen, X.; Chen, Q.; Zhang,
  B.; and Huang, W. 2026.
\newblock {OmniDrive-R1}: Reinforcement-Driven Interleaved Multi-Modal
  Chain-of-Thought for Trustworthy Vision-Language Autonomous Driving.
\newblock arXiv:2512.14044.

\bibitem[{Zheng et~al.(2025)Zheng, Mao, Ye, Li, Zhan, Lang, and
  Zhao}]{zhengDriveAgentR12025}
Zheng, W.; Mao, X.; Ye, N.; Li, P.; Zhan, K.; Lang, X.; and Zhao, H. 2025.
\newblock {DriveAgent-R1}: Advancing {VLM}-Based Autonomous Driving with Active
  Perception and Hybrid Thinking.
\newblock arXiv:2507.20879.

\bibitem[{Zhou et~al.(2025)Zhou, Liang, Tu, Chen, Ding, Zhang, Tan, Zhao, and
  Bai}]{zhouHERMES2025}
Zhou, X.; Liang, D.; Tu, S.; Chen, X.; Ding, Y.; Zhang, D.; Tan, F.; Zhao, H.;
  and Bai, X. 2025.
\newblock {HERMES}: A Unified Self-Driving World Model for Simultaneous {3D}
  Scene Understanding and Generation.
\newblock In \emph{Proceedings of the IEEE/CVF International Conference on
  Computer Vision (ICCV)}, 27817--27827.

\end{thebibliography}

\clearpage
\appendix
%
%
%

\section{Additional Method Details}
\label{app:method}

\subsection{Scene-Level Short-Term Memory Construction}
\label{app:mem_stm}
The scene-level short-term memory represents the current driving scene as a
compact graph containing the ego state, key objects, spatial sectors, distance
bands, motion states, risks, and inter-object relations. Adjacent-timestamp
multi-view images are passed through a pretrained HERMES-style BEV encoder and
an LLM decoder to obtain scene descriptions that initialize the graph-structured
state; The short-term memory is provided as a read-only scene prior during
each reasoning episode and is refreshed only when a new pair of adjacent
observations becomes available.

\subsection{Cross-Scene Long-Term Memory}
\label{app:mem_ltm}
\begin{algorithm}
	\caption{Offline Long-Term Memory Consolidation}
	\label{alg:consolidation}
	\begin{algorithmic}[1]
		\REQUIRE Current memory $M_l^{(n)}$, pending trajectory
		buffer $\mathcal{P}_n$, utility-update rate $\eta$
		\ENSURE Updated memory $M_l^{(n+1)}$
		
		\STATE $\mathcal{P}_n^{+}
		\gets
		\textsc{VerifiedTrajs}(\mathcal{P}_n)$
		
		\STATE $\mathcal{B}_n^{+}
		\gets
		\textsc{QualityFilter}
		\bigl(
		\textsc{ExtractBuffer}(\mathcal{P}_n^{+})
		\bigr)$
		
		\STATE $\mathcal{U}_n
		\gets
		\textsc{CollectUsefulEntries}(\mathcal{P}_n^{+})$
		
		\STATE $\widehat{M}_l^{(n)}
		\gets
		\textsc{Reinforce}
		\bigl(
		M_l^{(n)},\mathcal{U}_n,\eta
		\bigr)$
		
		\STATE $\{\mathcal{G}_k\}
		\gets
		\textsc{SemanticCluster}(\mathcal{B}_n^{+})$
		
		\FOR{each cluster $\mathcal{G}_k$}
		\STATE $\mathcal{G}_k
		\gets
		\textsc{Deduplicate}(\mathcal{G}_k)$
		
		\STATE $\mathcal{G}_k
		\gets
		\textsc{ResolveConflicts}
		\bigl(
		\mathcal{G}_k,\widehat{M}_l^{(n)}
		\bigr)$
		
		\STATE $\bar{e}_k
		\gets
		\textsc{Abstract}(\mathcal{G}_k)$
		\ENDFOR
		
		\STATE $M_l^{(n+1)}
		\gets
		\textsc{Merge}
		\bigl(
		\widehat{M}_l^{(n)},\{\bar{e}_k\}
		\bigr)$
		
		\STATE \textbf{return} $M_l^{(n+1)}$
	\end{algorithmic}
\end{algorithm}
The cross-scene long-term memory is built in two stages. It is first
initialized with $23{,}388$ experience entries of traffic rules and general
driving knowledge abstracted from the OmniDrive
dataset. After multi-step teacher rollout
(Appendix~\ref{app:rollout}), we further abstract reusable success patterns and
tool strategies from the verified trajectories and append them to the store.

Both the query and the stored entries are encoded into normalized
$384$-dimensional representations using
\texttt{sentence-transformers/all-MiniLM-L6-v2}. The retrieval query is
composed from the current question together with the scene-level short-term
memory, so that a single semantic query expresses both what the task asks and
what the current scene looks like; entries whose task and scene context both
resemble the query therefore rank highest. Retrieval computes cosine similarity
between the query and every entry, and the top-$4$ entries are returned.

Two safeguards apply at retrieval time. First, entries originating from the
query's own scene are excluded, which prevents a sample from retrieving the
roughly forty temporally adjacent frames of its own scene as trivially similar
evidence. Second, the returned view is inference-safe: internal identity fields
and the teacher's decision are stripped, so no raw training question or
ground-truth answer is ever exposed to the model. The store remains read-only
during benchmark evaluation and is updated only in the offline-consolidation
experiment (Algorithm~\ref{alg:consolidation}).

\subsection{Driving Tool Library}
\label{app:tools}

The complete driving tool library contains $27$ callable functions covering
visual grounding, metric geometry, object perception, ego-state estimation,
trajectory forecasting, occupancy reasoning, and map querying. Below, we
describe the $20$ core tools used in the reported experiments. Each tool
returns structured observations that can be incorporated into the transient
scene belief during memory-conditioned reasoning.

\paragraph{Open-Vocabulary Grounding Functions.}
\begin{itemize}\setlength{\itemsep}{2pt}
\item \textbf{detect\_objects\_2d\_open\_vocab}: Given a list of object words
and one named camera view, this function runs an open-vocabulary 2D detector and
returns bounding boxes, centers, and confidences. An empty detection result is treated as inconclusive rather than as evidence that the queried object is absent.
\item \textbf{select\_referred\_instance}: Selects and confirms the
referred instance when multiple same-class candidates are detected.
It takes the camera view, textual object description, and selected
bounding box as input, and returns the confirmed pixel center and box
for subsequent metric estimation.
\end{itemize}

\paragraph{3D Geometry Functions.}
\begin{itemize}\setlength{\itemsep}{2pt}
\item \textbf{estimate\_object\_3d\_location\_camera}: Given a camera view and
either a text query or an explicit box, this function returns the metric
camera-frame $(x,y,z)$ of the target in meters from multi-view BEV-3D
perception, together with the straight-line camera distance. The returned coordinates share a common metric BEV frame, enabling
cross-view spatial comparison.
\item \textbf{estimate\_object\_depth}: Returns the metric forward distance to a
target on one camera view, its straight-line camera distance, and a coarse
distance band. Used for ``how far'' questions.
\item \textbf{measure\_pairwise\_distance\_3d}: Given two targets on one camera
view, returns the absolute lateral gap $|\Delta x|$ and the straight-line 3D
distance between them in meters.
\item \textbf{compare\_object\_depth\_order}: Given two targets on one camera
view, reports which is closer to the ego, their metric depths, and a confidence
level. Used for ``which is closer'' and front/behind questions.
\item 
\textbf{estimate\_object\_depth\_monocular}: Provides a
monocular-depth estimation. The result is treated as an
approximate pixel-anchored estimate and is considered less reliable
for long-range targets. For pairwise measurement, both targets are
estimated within the same image to preserve relative scale
consistency.
\end{itemize}

\paragraph{Detection Functions.}
\begin{itemize}\setlength{\itemsep}{2pt}
\item \textbf{get\_leading\_object\_detection}: Detects the leading object in
the same lane within $10$\,m ahead, returning its ID, position, and size.
Returns None if no leading object exists.
\item \textbf{get\_surrounding\_object\_detections}: Detects objects within a
$20\,\mathrm{m}\times20\,\mathrm{m}$ box around the ego, returning IDs,
positions, and sizes.
\item \textbf{get\_front\_object\_detections}: Identifies objects in a
$10\,\mathrm{m}\times20\,\mathrm{m}$ region in front of the ego, returning IDs,
positions, and sizes.
\item \textbf{get\_object\_detections\_in\_range}: Detects objects whose center
falls within a specified BEV rectangle
$(x_{\mathrm{start}},x_{\mathrm{end}})\times(y_{\mathrm{start}},y_{\mathrm{end}})$
in meters.
\end{itemize}

\paragraph{Ego-State Estimation Function.}
\begin{itemize}\setlength{\itemsep}{2pt}
\item \textbf{get\_ego\_states}: Returns the ego vehicle's current kinematic
state and short history: velocity $(v_x,v_y)$, heading angular velocity,
acceleration, can-bus position, heading speed, steering angle, the last two
seconds of trajectory, and the mission goal (LEFT / RIGHT / FORWARD).
\end{itemize}

\paragraph{Trajectory Forecasting Functions.}
\begin{itemize}\setlength{\itemsep}{2pt}
\item \textbf{get\_future\_trajectories\_for\_specific\_objects}: Given a list
of object IDs obtained from a prior detection call, returns their predicted
future waypoints over the next $\sim\!3$\,s.
\item \textbf{get\_future\_trajectories\_in\_range}: Returns predicted future
trajectories for any object whose current center lies in a specified BEV
rectangle.
\end{itemize}

\paragraph{Occupancy and Map Query Functions.}
\begin{itemize}\setlength{\itemsep}{2pt}
\item \textbf{get\_occupancy\_at\_locations\_for\_timestep}: Given a list of
$(x,y)$ locations and a future timestep $t\in\{0,1,2,3,4\}$ (corresponding to
$0$, $0.5$, $1$, $1.5$, $2$\,s), returns whether each location is occupied.
\item \textbf{get\_drivable\_at\_locations}: Returns whether each $(x,y)$
location lies in a drivable region.
\item \textbf{get\_lane\_category\_at\_locations}: Returns the lane-element
category at each $(x,y)$ location, one or more of \{divider, ped\_crossing,
boundary\}, with optional probability scores.
\item \textbf{get\_current\_shoulder}: Returns the distance from the ego to the
left and right road shoulders at the current location.
\item \textbf{get\_current\_lane\_divider}: Returns the distance from the ego to
the left and right lane dividers at the current location.
\item \textbf{get\_nearest\_pedestrian\_crossing}: Returns the location of the
nearest pedestrian crossing in front of the ego, or None if none exists.
\end{itemize}

\section{Data Generation and Post-Training Details}
\label{app:data}

\subsection{Multi-Step Teacher Rollout and Validation}
\label{app:rollout}
We construct gold-conditioned, tool-executing teacher trajectories.
The gold answer is provided to the teacher to anchor trajectory
synthesis, while real tool observations are obtained only through
execution. Validation therefore focuses on action validity,
observation provenance, grounding quality, and
observation--reasoning consistency rather than independently
re-evaluating answer correctness. Each rollout follows the flow:
\begin{quote}\small
	$q + I + M_s + \mathcal{E}_K \rightarrow$ Reasoning State $1 \rightarrow$ Tool
	Call $1 \rightarrow$ Observation $1 \rightarrow \cdots \rightarrow$ Final Answer,
\end{quote}
with real tool execution at each step. Validation applies action-syntax and
tool-argument checks, object-grounding and observation-provenance checks,
observation--reasoning consistency, final-answer correctness, and
maximum-step / redundancy filtering; failed rollouts are regenerated. 

Table~\ref{tab:teacher_cfg} records the exact teacher setup, so that the
generation run is documented even though it cannot be replayed bit-exactly.
The model identifier is a provider-side \emph{alias} rather than a dated
snapshot: the API returns no finer build string, so the alias may resolve to a
different underlying build after our call window. We therefore treat the
released trajectories as the frozen record of this run rather than as an
artifact that a reader can regenerate. Extended-thinking mode was not enabled;
the teacher emits its trace directly under a JSON response-format constraint.
Nucleus and top-$k$ sampling parameters were left at provider defaults and are
not part of the recorded configuration.

\begin{table}[t]
	\centering\small
	\setlength{\tabcolsep}{3pt}
	\begin{tabular}{@{}l>{\raggedright\arraybackslash}p{3.4cm}@{}}
		\toprule
		Item & Value\\
		\midrule
		API provider          & Alibaba Cloud DashScope\\
		Model identifier      & \texttt{qwen3.6-plus} (alias)\\
		Dated snapshot        & not exposed by the API\\
		Extended thinking     & disabled\\
		Call window           & 2026-05-09 to 2026-05-13\\
		Logged calls          & 6{,}395\\
		\midrule
		Temperature           & 0.1\\
		Top-$p$ / top-$k$     & provider defaults (not set)\\
		Max output tokens     & 2{,}048\\
		Response format       & JSON object mode\\
		Request timeout       & 180\,s\\
		\midrule
		Retries per question  & 2 (3 attempts max)\\
		\bottomrule
	\end{tabular}
	\caption{Teacher generation configuration for Phase-1 trajectory synthesis.}
	\label{tab:teacher_cfg}
\end{table}

The teacher system prompt and the tool-call JSON schema it must satisfy are
reproduced in Appendix~\ref{app:teacher_prompts}; the schema is the same
registry-validated contract used at inference (Appendix~\ref{app:tools}), so a
trajectory that passes validation is executable by the student without
translation.

\subsection{Stepwise Conversion and Two-Phase SFT}
\label{app:sft}
The verified trajectory pool is decomposed into step-level supervision rather
than used as whole traces, which are too open-ended for supervised fine-tuning.
Each trajectory yields several next-step prediction examples: from the question
plus compact memory context to $\langle$think$\rangle$ and
$\langle$action$\rangle$; from the previous observation to the next
$\langle$think$\rangle$ and $\langle$action$\rangle$; and from the final
observation to $\langle$think$\rangle$, reasoning, and the final answer.
Environment-returned observations appear as user-side turns and are masked from
the loss, so supervision applies only to model-generated reasoning, actions,
belief updates, and final answers. Phase~1 yields $13{,}679$ tool-decision
records and Phase~2 yields $20{,}062$ records with complete memory-conditioned
reasoning contexts and executed tool observations.

We adapt Qwen2.5-VL-7B-Instruct with LoRA (rank $16$, $\alpha=32$, dropout
$0.05$) on all linear layers of the language backbone, freezing the vision
encoder and multimodal aligner; all stages use bfloat16 and AdamW. Phase~1
learns valid tool selection and argument generation, with tool observations
removed from the context except candidate detections needed by the
instance-selection action. Phase~2 continues from the Phase~1 adapter and learns
observation-grounded reasoning. The complete training, optimization,
LoRA, and data-selection hyperparameters for both SFT phases are
summarized in Table~\ref{tab:sft_hparams}.

\begin{table}[t]
	\centering\small
	\setlength{\tabcolsep}{4pt}
	\begin{tabular}{@{}lcc@{}}
		\toprule
		Parameter & Phase-1 SFT & Phase-2 SFT\\
		\midrule
		\multicolumn{3}{@{}l}{\emph{Training Configuration}}\\
		\quad Objective               & tool-call init & obs.-grounded reason.\\
		\quad Epochs                  & 3 & 20\\
		\quad Per-device batch size   & 1 & 1\\
		\quad Gradient accumulation   & 16 & 16\\
		\quad Data-parallel size      & 1 & 1\\
		\quad \textbf{Effective batch size} & \textbf{16} & \textbf{16}\\
		\quad Max sequence length     & 12{,}288 & 12{,}288\\
		\quad Precision               & bfloat16 & bfloat16\\
		\quad Attention               & SDPA & SDPA\\
		\quad Gradient checkpointing  & enabled & enabled\\
		\quad Loss scope              & last round & last round\\
		\midrule
		\multicolumn{3}{@{}l}{\emph{Optimizer (AdamW, fused)}}\\
		\quad Learning rate           & 5e-5 & 2e-5\\
		\quad Betas $(\beta_1,\beta_2)$ & (0.9, 0.95) & (0.9, 0.95)\\
		\quad Epsilon $\epsilon$      & 1e-8 & 1e-8\\
		\quad Weight decay            & 0.01 & 0.01\\
		\quad Gradient clipping       & 1.0 & 1.0\\
		\midrule
		\multicolumn{3}{@{}l}{\emph{Scheduler}}\\
		\quad Type                    & Cosine & Cosine\\
		\quad Warmup ratio            & 0.05 & 0.05\\
		\midrule
		\multicolumn{3}{@{}l}{\emph{LoRA Configuration}}\\
		\quad Rank                    & 16 & 16\\
		\quad Alpha                   & 32 & 32\\
		\quad Dropout                 & 0.05 & 0.05\\
		\quad Bias                    & none & none\\
		\quad Target modules          & \multicolumn{2}{c}{all linear layers}\\
		\quad Frozen                  & \multicolumn{2}{c}{vision encoder, aligner}\\
		\midrule
		\multicolumn{3}{@{}l}{\emph{Data and Selection}}\\
		\quad Training records        & 13{,}679 & 20{,}062\\
		\quad Validation split ratio  & 0.01 & 0.005\\
		\quad Data-loader workers     & 4 & 4\\
		\quad Save / eval interval    & 200 / 300 & 1000 / 1000\\
		\quad Selection criterion     & \multicolumn{2}{c}{validation loss}\\
		\quad Seed                    & 42 & 42\\
		\bottomrule
	\end{tabular}
	\caption{Supervised fine-tuning hyperparameters. Both phases train a single
		LoRA adapter on one GPU; the effective batch size is
		$1 \times 16 \times 1 = 16$ sequences.}
	\label{tab:sft_hparams}
\end{table}

\subsection{GRPO and Reward Design}
\label{app:grpo}
Starting from the Phase~2 checkpoint, we run trajectory-level GRPO with lr
$1\times10^{-6}$, warmup ratio $0.25$, KL coefficient $\beta=0.001$, and $G=8$
sampled trajectories per prompt; the maximum completion length is $1{,}536$
tokens and training uses DeepSpeed ZeRO-3. A single rule-based outcome reward
model assigns each sampled trajectory to exactly one tier
(Table~\ref{tab:reward}). Correctness is evaluated on the multiple-choice letter
of the last parseable final answer; \emph{tool} denotes at least one action
passing the schema legality check; \emph{memory cited} requires at least two
concrete scene-state or experience phrases surfaced in the reasoning. Three properties of the reward deserve explicit statement:

\begin{table}
	\centering
	\setlength{\tabcolsep}{3pt}
	\begin{tabular}{clc}
		\toprule
		Tier & Condition & Reward\\
		\midrule
		1 & correct $\wedge$ tool $\wedge$ memory cited & $+0.7$\\
		2 & correct $\wedge$ tool $\wedge$ no memory citation & $+0.5$\\
		3 & correct $\wedge$ no tool & $0.0$\\
		4 & incorrect $\wedge$ tool & $+0.2$\\
		5 & incorrect $\wedge$ no tool & $0.0$\\
		6 & malformed output & $0.0$\\
		\bottomrule
	\end{tabular}
	\caption{Tiered outcome reward for trajectory-level GRPO. Tiers are mutually
	exclusive; each trajectory is scored once.}
	\label{tab:reward}
\end{table}

\begin{table}[t]
	\centering\small
	\setlength{\tabcolsep}{3pt}
	\begin{tabular}{@{}l>{\raggedright\arraybackslash}p{3.05cm}@{}}
		\toprule
		Parameter & Value\\
		\midrule
		\multicolumn{2}{@{}l}{\emph{Objective}}\\
		\quad Loss / advantage estimator & GRPO\\
		\quad Clipping parameter $\epsilon$   & 0.2 (symmetric)\\
		\quad KL coefficient $\beta$          & 0.001\\
		\quad KL placement                    & in loss, not in reward\\
		\quad KL estimator                    & sampled-token ($k_3$)\\
		\quad Importance sampling             & token level\\
		\quad Policy update epochs $\mu$      & 1 (on-policy)\\
		\quad Entropy bonus                   & none\\
		\midrule
		\multicolumn{2}{@{}l}{\emph{Sampling}}\\
		\quad Group size $G$                  & 8\\
		\quad Generation batch size           & 16\\
		\quad Prompts per gen.\ batch         & 2\\
		\quad Steps per generation            & 8\\
		\quad Temp.\ / top-$p$ / top-$k$      & 1.0 / 0.95 / 50\\
		\quad Max prompt / completion         & 512 / 1{,}536\\
		\quad Max tool-interaction turns      & 7\\
		\quad Rollout engine                  & vLLM (server mode)\\
		\midrule
		\multicolumn{2}{@{}l}{\emph{Reward}}\\
		\quad Reward function                 & \texttt{answer\_\allowbreak or\_tool}\\
		\quad Reward weight                   & 1.0\\
		\quad Standardization                 & per group\\
		\quad Reward whitening                & disabled\\
		\quad Overlong / dyn.\ filtering      & disabled\\
		\midrule
		\multicolumn{2}{@{}l}{\emph{Reference policy}}\\
		\quad Source                          & Phase-2 SFT adapter\\
		\quad Synchronization                 & none (frozen)\\
		\midrule
		\multicolumn{2}{@{}l}{\emph{Optimization}}\\
		\quad Per-device batch size           & 2\\
		\quad Gradient accumulation           & 8 (4 in final phase)\\
		\quad \textbf{Effective batch size}   & \textbf{16}\\
		\quad Optimizer                       & AdamW (fused)\\
		\quad Betas / $\epsilon$ / wd         & (0.9, 0.95) / 1e-8 / 0.01\\
		\quad Gradient clipping               & 1.0\\
		\quad Scheduler                       & Cosine\\
		\quad Parallelism                     & ZeRO-3; 1 trainer $+$ 1 rollout GPU\\
		\quad Save interval                   & every 50 steps\\
		\quad Seed / data seed                & 42 / 42\\
		\bottomrule
	\end{tabular}
	\caption{GRPO implementation parameters shared across the four continued-training
		phases. Per-phase learning rate, warmup ratio, and step budget are listed in
		Table~\ref{tab:grpo_chain}.}
	\label{tab:grpo_hparams}
\end{table}

\begin{table}[t]
	\centering\small
	\setlength{\tabcolsep}{4pt}
	\begin{tabular}{@{}lcccc@{}}
		\toprule
		Phase & LTM & LR & Warmup & Steps\\
		\midrule
		v0 (fresh) & ---     & $1\!\times\!10^{-6}$ & 0.25 & 800\\
		v1         & ---     & $5\!\times\!10^{-7}$ & 0.05 & 500\\
		v2         & enabled & $5\!\times\!10^{-7}$ & 0.10 & 500\\
		v3 (final) & enabled & $1\!\times\!10^{-6}$ & 0.10 & 500\\
		\bottomrule
	\end{tabular}
	\caption{GRPO continued-training chain. Each phase resumes strictly from the
	preceding phase's checkpoint; long-term memory is injected into the prompt
	set from v2 onward. The reported model is the final v3 checkpoint.}
	\label{tab:grpo_chain}
\end{table}

First, the memory-citation bonus is fully correctness-gated: the $+0.2$ gap
between Tier~1 and Tier~2 is available only on correct answers.

Second, tool use is deliberately not fully correctness-gated. Tier~4 assigns a
small positive reward to incorrect trajectories that issued at least one legal
tool call, which makes $R(\text{incorrect},\text{tool}) > R(\text{correct},
\text{no tool})$. This ordering is intentional and serves two purposes. It is a
bootstrapping term: at the start of reinforcement training the policy must first
learn that the tool interface is executable at all, and a purely
correctness-gated reward gives no gradient on the prompts the model cannot yet
answer, which are exactly the prompts tool use is meant to address. It is also
an anti-shortcut term: the supervised warm-up already leaves the model able to
answer many questions directly, so rewarding correct-but-toolless trajectories
would make direct answering the dominant strategy and discard the interaction
behaviour that the warm-up established.

The shaping term cannot be exploited, for three reasons that are structural
rather than empirical. (i)~Tiers are mutually exclusive and each trajectory is
scored exactly once, so a trajectory that issues seven tool calls receives the
same $+0.2$ as one that issues a single call; the reward is therefore flat in
the number of calls and exerts no pressure toward deeper or repeated
invocation. (ii)~The tier is strictly dominated: any trajectory that converts a
tool call into a correct answer moves to Tier~2 or Tier~1 and more than doubles
its reward, so the shaping term is never the best available outcome for a
prompt the model can solve. (iii)~Advantages are standardized within each group
of $G$ samples of the same prompt, so Tier~4 only produces a positive advantage
when no sample in the group is correct; whenever the group contains a correct
trajectory, the tool-using-but-incorrect samples receive negative advantage.
The shaping term is thus active precisely on the prompts where the policy has
not yet learned to succeed, and inactive elsewhere.

Third, open-ended items are treated as incorrect by the multiple-choice
correctness test and can therefore only reach Tiers~4--6, so the reinforcement
stage applies pressure on multiple-choice correctness and tool discipline but no
direct pressure on open-ended reasoning quality. We return to this asymmetry in
Appendix~\ref{app:stats}.

\subsection{Training and Compute Configuration}
\label{app:compute}
GRPO uses a separate subset of training questions disjoint from SFT. At
inference the model follows a Think--Action--Observation--Update loop
with real tool execution: each returned observation is appended to the reasoning
context and informs the next step, but does not modify the explicit short-term
memory. We use greedy decoding (temperature $0$), bfloat16, SDPA attention, an
image-pixel budget of $\sim\!10^{6}$, and we allow at most seven tool-execution turns followed by one terminal answer turn, yielding at most eight model turns per query. Unless otherwise stated, all model parameters and both memory channels
are frozen during evaluation. All experiments run on $4\times$ NVIDIA RTX PRO
6000 Blackwell GPUs ($96$\,GB each) distributed across two servers.

\paragraph{Software environment.}
Ubuntu 24.04.2 LTS (kernel 6.8.0), NVIDIA driver 590.48.01, CUDA 12.8 with
cuDNN 9.10.2, Python 3.10. Key packages: torch 2.8.0 (cu128), transformers
4.57.6, peft 0.18.1, trl 0.24.0, deepspeed 0.17.6, accelerate 1.13.0, vllm
0.11.0, sentence-transformers 5.4.1, numpy 2.2.6. \texttt{flash-attn} is
deliberately not installed and attention runs through SDPA; environments that do
install it may produce slightly different numerics. A complete \texttt{pip
freeze} of this environment is included in the supplement.

\paragraph{Random seeds.}
Trainer-side randomness is seeded at $42$ and propagated to Python, NumPy, and
PyTorch, fixing adapter initialization, data shuffling, and dataloader order for
both supervised stages and the reinforcement stage. Dataset partitions used for
analysis are drawn under separately recorded seeds. Inference is deterministic
given a fixed checkpoint. The teacher model and the evaluation judge are hosted third-party APIs that accept no seed argument, so the released
trajectories and judge outputs are frozen records of the actual runs; and
reinforcement rollouts are sampled at temperature $1$ without a per-sample seed
threaded through the rollout server, so the sampled groups are not
bit-reproducible across reruns, as is expected for on-policy reinforcement
learning.

\paragraph{Hyperparameter selection.}
Optimizer, adapter, and reinforcement hyperparameters were fixed by compute
budget rather than searched: they follow the defaults of the training toolkit
and of prior stepwise tool-use recipes. We ran no grid or random search over
learning rate, adapter rank, retrieval depth, group size, or interaction budget,
and we state this plainly rather than present a post-hoc search space.
Checkpoints are selected by accuracy on a validation partition held out from the
\emph{training} file by the trainer, never on any benchmark test split.

\section{Benchmarks and Evaluation Protocols}
\label{app:benchmarks}

\subsection{DriveLMM-o1 Protocol}
\label{app:drivelmm}
\textbf{DriveLMM-o1}~\citep{ishaqDriveLMMo1StepbystepReasoning2025} is our
primary benchmark for reasoning quality and answer accuracy. It is built on
nuScenes keyframes and contains over $18$K training and over $4$K test VQA
examples spanning perception, prediction, and planning; unlike earlier driving
VQA sets it supplies manually curated intermediate reasoning steps for every
question rather than only a final answer. The test split contains $4{,}634$
questions ($2{,}391$ multiple-choice and $2{,}243$ open-ended), drawn from $539$
multi-view keyframes across $115$ nuScenes scenes; each sample provides six
surround-view images.

\paragraph{Input configuration.}
We follow the official inference format. The six views are supplied in the
canonical order (FRONT\_LEFT, FRONT, FRONT\_RIGHT, BACK\_RIGHT, BACK,
BACK\_LEFT), stitched into a $2\times3$ grid, and the layout is declared in the
system prompt so that answers referring to, for example, ``top row, second
image'' resolve correctly. We reuse the official reasoning instruction verbatim
so that the fine-tuned model and the evaluation harness see the same template.

\paragraph{Reasoning score.}
An LLM judge scores each response against the human-verified reasoning steps on
a $1$--$10$ scale along twelve dimensions: Faithfulness-Step,
Informativeness-Step, Risk Assessment Accuracy, Traffic Rule Adherence, Scene
Awareness \& Object Understanding, Repetition-Token, Hallucination, Semantic
Coverage-Step, Commonsense Reasoning, Missing Step, Relevance, and Missing
Details. The judge receives the question, the concatenated ground-truth
reasoning steps and final answer, and the model response, and returns a
structured JSON object under a fixed schema with no free text. The full prompts given to the judge is presented in~\ref{app:judge_prompt}. We use
GPT-4o-mini at temperature $0$, matching the reference implementation's default.
The reported reasoning score is
\begin{equation}
\mathrm{Reason.}
= \frac{100}{10\,|\mathcal{D}|}\sum_{i\in\mathcal{D}}
  \frac{1}{12}\sum_{k=1}^{12} m_{k,i},
\label{eq:reason}
\end{equation}
where $m_{k,i}$ is dimension $k$ for question $i$ and $\mathcal{D}$ is the full
$4{,}634$-question split. Table~\ref{tab:drivelmmo1} reports five of the twelve
dimensions individually; the ``Missing'' column is Missing Details.

\paragraph{Answer accuracy.}
A question is multiple-choice when the trailing index of its identifier lies in
$\{2,3,4,5,8\}$, the official partition, which yields exactly $2{,}391$ items.
The predicted option is extracted by splitting the response at the first
final-answer marker and taking the first option pattern in the remainder; a
response with no parseable marker is scored incorrect. MCQ accuracy is the mean
over the $2{,}391$ items.

\subsection{Cross-Benchmark Evaluation}
\label{app:cross_bench}
\textbf{DriveMLLM}~\citep{guoDriveMLLMBenchmark2024} evaluates fine-grained
spatial understanding on nuScenes front-facing images, pairing each image with
programmatically generated questions whose ground truth is derived from
calibrated 3D boxes and camera intrinsics. It covers both absolute
(camera-to-object) and relative (object-to-object) relations, and each question
is answered from a single view with no temporal context. Following the zero- and
one-shot protocols of prior tool-augmented work, we evaluate eight tasks:
left/right (L/R), front/behind (F/B), relative horizontal distance (RHD),
relative distance (RD), pixel localization (PPos), bounding-box localization
(BBox), camera vertical distance (CVD), and camera distance (CD). Binary
relational tasks are scored by exact match,
$\mathrm{acc}_i = \mathbf{1}[p_i = y_i]$. Metric-regression tasks use a bounded
penalty on the absolute error,
\begin{equation}
\mathrm{acc}_i = \frac{1}{1+\alpha_d\,\lVert d_i - d_i^{\mathrm{gt}}\rVert_1},
\qquad \alpha_d = 0.05,
\end{equation}
covering RHD, RD, CVD, and CD. Pixel localization uses the Euclidean form
\begin{equation}
\mathrm{acc}_i = \frac{1}{1+\alpha_p\,\lVert x_i - x_i^{\mathrm{gt}}\rVert_2},
\qquad \alpha_p = 0.005,
\end{equation}
and bounding-box localization is scored by intersection-over-union. The
aggregate score is the unweighted mean over the eight tasks,
$\mathrm{AccS} = \frac{1}{8}\sum_{j=1}^{8}\mathrm{acc}_j$. All scaling constants
are the published values. Parameters and long-term memory stay frozen; the
one-shot setting adds a single in-context demonstration.

\textbf{STSBench}~\citep{fruhwirthSTSBench2025} targets spatio-temporal
reasoning over both ego and non-ego agents, which distinguishes it from
benchmarks that test ego-centric action recognition on single images or
monocular video. Its nuScenes instantiation mines $43$ pre-defined traffic
scenarios from ground-truth annotations, passes them through human verification,
and converts them into $971$ five-way multiple-choice questions, each grounded in
six temporal frames at $2$\,Hz from six surround-view cameras. We report accuracy
for ego motion, ego--agent relations, agent motion, and agent--agent relations,
plus a micro-average over all $971$ questions.

\textbf{STRIDE-QA}~\citep{ishiharaSTRIDEQA2026} is built from an urban driving
corpus disjoint from nuScenes, which makes it a genuine cross-dataset test:
neither our training data nor our memory contains any STRIDE-QA scene, sensor
configuration, or city. We evaluate on the benchmark split, built from held-out
recording dates and comprising $5{,}317$ QA pairs over $409$ scene groups,
filtered to dynamic interactions and categorized into six scenario types. The
model observes four context frames from a front-facing camera with a
$60^{\circ}$ field of view at $t\in\{-1.5,-1.0,-0.5,0\}$\,s, with a single target
agent identified by its segmentation mask, and predicts at
$t\in\{0,1,2,3\}$\,s the target's distance, velocity, and heading angle together
with the ego velocity. Heading is expressed in the ego frame with $0^{\circ}$
forward and positive counter-clockwise; targets may leave the field of view at
$t>0$. A localization succeeds only if distance and heading are simultaneously
within tolerance,
\begin{equation}
s_{g,t} = \mathbf{1}\!\left[
\lvert \hat{d}_t - d_t^{*}\rvert < 0.25\,d_t^{*}
\;\wedge\;
\lvert \hat{\theta}_t - \theta_t^{*}\rvert < 10^{\circ}
\right],
\end{equation}
where the $\pm25\%$ distance margin follows prior spatial-reasoning work and the
$\pm10^{\circ}$ heading margin is set so that at $10$\,m the lateral deviation
equals a standard $3.5$\,m lane width. We report LSR at the $0/1/2/3$\,s
timesteps, its mean
$\mathrm{MLSR}=\frac{1}{|G|}\sum_{g}\frac{1}{T+1}\sum_{t=0}^{T}s_{g,t}$,
and Temporal Localization Consistency
$\mathrm{TLC}=\frac{1}{|G|}\sum_{g}\mathbf{1}[s_{g,0}\wedge s_{g,1}\wedge s_{g,2}\wedge s_{g,3}]$
with $T=3$, the strict fraction of sequences localized correctly at all four
timesteps.

\subsection{Leakage Control}
\label{app:metrics}

\paragraph{Training data.}
All supervised trajectories and all reinforcement prompts are generated from the
DriveLMM-o1 \emph{training} split only. No test question, image, or reference
answer is used at any training stage.

\paragraph{Memory initialization.}
The initial long-term store is abstracted from the OmniDrive
dataset, which supplies scene-level driving knowledge,
traffic rules, and counterfactual driving rationales over nuScenes scenes.
Because OmniDrive shares its underlying corpus with three of our benchmarks,
extraction is restricted to nuScenes \emph{training} scenes: no entry in the
store originates from a scene that appears in any evaluation split. Entries
added later, during the abstraction of reusable success and failure patterns
from verified teacher rollouts, inherit the same restriction because those
rollouts are themselves generated only from training-split questions.

\paragraph{Memory access at evaluation time.}
During benchmark evaluation the store is read-only and frozen. Retrieval applies
two safeguards. First, an anti-leak filter excludes every entry whose scene
identifier matches the query's own scene, which prevents a sample from
retrieving the roughly forty temporally adjacent frames of its own scene as
trivially similar evidence. Second, the returned view is inference-safe: teacher
decisions and dataset identity tokens are stripped, so no reference answer or
scene identifier is ever exposed to the model.

\paragraph{Model selection.}
Checkpoints are selected on a validation partition held out from the training
file by the trainer, never on any benchmark test split.

\paragraph{Offline consolidation.}
Consolidation uses only training-split rollouts and additional non-benchmark
scenes; correctness is verified against training labels alone, and the memory is
frozen before evaluation, so the self-evolution protocol is strictly inductive.

\section{Prompt and Serialization Templates}
\label{app:prompts}

\subsection{Teacher Rollout Prompts}
\label{app:teacher_prompts}
Teacher trajectories are produced by a three-stage prompted pipeline rather
than a single generation call. 
Stage~1 plans a short action skeleton without
access to any tool output. 
Stage~2 is invoked once per interaction round: it
observes the tool results executed so far and either emits the next action or
declares the evidence sufficient. 
Stage~3 receives the completed trace together
with the real observations, then updates current belief and step by step reasoning. All three stages are conditioned on the
gold answer, so a trajectory is an \emph{explanation} of a known answer rather
than an independent attempt at it; the filtering described in
Appendix~\ref{app:rollout} therefore rejects malformed or ungrounded traces
rather than incorrect ones. The three system prompts are reproduced verbatim
below.

\paragraph{Stage 1 --- action planning.}
\emph{You are a Memory-Tool Synergistic Reasoning policy teacher. Your job is to produce a
	skeleton for autonomous driving VQA. The skeleton must decide
	which external actions should be executed before the student answers. You see
	the image grid, question, gold answer, memory and reasoning seed, but you DO NOT see
	real tool observations. Do not fabricate observations.}

\emph{Output strict JSON only:}
\begin{quote}\small\ttfamily
	\{"steps": [\{"think": \{"goal": "...", \\
	"evidence\_gap": "..."\}, \\
	"action": \{"type": "...", "arguments": \{...\}\}\}]\}
\end{quote}

\emph{Rules:}
\begin{enumerate}\small\setlength{\itemsep}{0pt}
	\item Use 1--3 actions only.
	\item \texttt{action.type} must be one of the allowed action types.
	\item Use \texttt{sample\_token="<CURRENT\_SAMPLE>"} when an action needs the
	current sample.
	\item Look up in short-term memory before calling tools.
	\item No \texttt{observation\_summary}, \texttt{final\_answer}, or tool result text in this phase.
\end{enumerate}

\paragraph{Stage 2 --- next-step decision.}
\emph{You are a Memory-Tool Synergistic Reasoning policy teacher. Decide the next action after reading the image, question, gold answer, reasoning
	seed, and any already executed observations. Use a memory-tool synergistic
	style: first use scene-level short-term memory when scene continuity is relevant,
	retrieve long-term experience when planning or safety judgment benefits from
	similar past driving cases, then use tools as needed.}

\emph{Output strict JSON only:}
\begin{quote}\small\ttfamily
	\{"decision\_type": "action", \\
	"scene\_state": "<brief reusable scene-local state>", \\
	"experience": "<brief retrieved cross-scene experience>", \\
	"think": \{"goal": "...", "evidence\_gap": "...", \\
	\ \ "decision": "<why this action is the next \\
	\ \ best evidence source>"\}, \\
	"action": \{"type": "...", "arguments": \{...\}\}\}
\end{quote}

\emph{If enough evidence has already been gathered and at least one action has
	been executed, output strict JSON only:}
\begin{quote}\small\ttfamily
	\{"decision\_type": "finish", \\
	"scene\_state": "...", "experience": "...", \\
	"think": \{"goal": "finish with the benchmark- \\
	\ \ compatible answer", "evidence\_gap": "none", \\
	\ \ "decision": "<why the evidence is sufficient>"\}, \\
	"final\_answer": "<exact gold final answer>"\}
\end{quote}

\emph{Rules:}
\begin{enumerate}\small\setlength{\itemsep}{0pt}
	\item Use at most one action in this response.
	\item \texttt{action.type} must be one of the available actions.
	\item Use \texttt{sample\_token="<CURRENT\_SAMPLE>"} when an action needs the
	current sample.
	\item For planning, safety, behavior prediction, relative motion, and ambiguous temporal questions, consider looking up in short-term memory first to get a full view of the scene.
	\item Avoid template collapse: do not repeat a fixed route just because the
	question type is familiar. Compare the current observations, memory
	availability, action history, tool costs, and unresolved evidence gap..
	\item \texttt{decision\_type=finish} is allowed only after at least one useful
	action has been executed.
	\item \texttt{final\_answer} in a finish response must exactly match the gold
	final answer.
\end{enumerate}

\paragraph{Stage 3 --- summary.}
\emph{You are a Memory-Tool Synergistic Reasoning policy teacher. You will
	receive a think-action- trace with real observations from tools and
	memory. Produce compact memory sections, observation-aware private updates, and
	DriveLMM-o1-style public step-by-step reasoning that turns private evidence into
	benchmark-friendly reasoning.}

\emph{Output strict JSON only:}
\begin{quote}\small\ttfamily
	\{"scene\_state": "<compact structured scene state>" \\
	"experience": "<brief historical driving experience>" \\
	"updates": [\{"observation\_summary": "...", \\
	\ \ "belief\_update": "...", "decision": "..."\}], \\
	"reasoning": "<step-by-step reasoning>", \\
	"final\_answer": "<exact gold final answer>"\}
\end{quote}

\emph{Rules:}
\begin{enumerate}\small\setlength{\itemsep}{0pt}
	\item \texttt{final\_answer} must exactly equal the provided gold final answer.
	\item \texttt{updates} length must equal the number of observations.
	\item \texttt{observation\_summary} must accurately reflect the given
	observation; do not add unsupported facts.
	\item \texttt{experience} must only use retrieved long-term experience context.
	\item \texttt{belief\_update} must be grounded in the corresponding observation.
	\item \texttt{reasoning} must be step-by-step style. Step~1 must point to key evidence in the
	$2\times3$ image grid or a camera view or spatial region, and include the real
	visual or observation evidence that supports it, without evidence source tags.
	Step~2 must explain how that evidence affects the ego vehicle's safety,
	maneuver feasibility, collision risk, lane-change feasibility, speed choice, or
	future behavior. 
	Step~3 must first name compact evidence sources, translate their support into natural
	driving language; for multiple-choice questions it must explain why the selected
	option is safer, more likely, or more relevant than the alternatives. Step~4,
	give a final answer.
\end{enumerate}

\subsection{LLM-Judge Prompt for DriveLMM-o1}
\label{app:judge_prompt}
We use the official DriveLMM-o1 evaluation script without modification. The
judge is \texttt{gpt-4o-mini} at temperature $0$, constrained to a strict JSON
schema, with a $500$-token output budget. For each question the judge receives a
single user message of the form
\begin{quote}\small\ttfamily
	Question: \{question\}\\
	Ground Truth: \{ground\_truth\}\\
	LLM Response: \{llm\_response\}
\end{quote}
where \texttt{ground\_truth} concatenates the human-written reasoning steps and
the reference final answer. The system prompt is reproduced verbatim below.

\paragraph{System prompt.}
\emph{You are an autonomous driving reasoning evaluator. Your task is to assess
	the alignment, coherence, and quality of reasoning steps in text responses for
	safety-critical driving scenarios.}

\emph{You will evaluate the model-generated reasoning using the following
	metrics:}

\begin{enumerate}\small\setlength{\itemsep}{1pt}
	\item \textbf{Faithfulness-Step (1--10)}: Measures how well the model's
	reasoning steps align with the ground truth.
	9--10: All steps correctly match or closely reflect the reference.
	7--8: Most steps align, with minor deviations.
	5--6: Some steps align, but several are incorrect or missing.
	3--4: Few steps align; most are inaccurate or missing.
	1--2: Majority of steps are incorrect.
	
	\item \textbf{Informativeness-Step (1--10)}: Measures completeness of reasoning.
	9--10: Captures almost all critical information.
	7--8: Covers most key points, with minor omissions.
	5--6: Missing significant details.
	3--4: Only partial reasoning present.
	1--2: Poor extraction of relevant reasoning.
	
	\item \textbf{Risk Assessment Accuracy (1--10)}: Evaluates if the model
	correctly prioritizes high-risk objects or scenarios.
	9--10: Correctly identifies and prioritizes key dangers.
	7--8: Mostly accurate, with minor misprioritizations.
	5--6: Some important risks are overlooked.
	3--4: Significant misjudgments in risk prioritization.
	1--2: Misidentifies key risks or misses them entirely.
	
	\item \textbf{Traffic Rule Adherence (1--10)}: Evaluates whether the response
	follows traffic laws and driving best practices.
	9--10: Fully compliant with legal and safe driving practices.
	7--8: Minor deviations, but mostly correct.
	5--6: Some inaccuracies in legal/safe driving recommendations.
	3--4: Several rule violations or unsafe suggestions.
	1--2: Promotes highly unsafe driving behavior.
	
	\item \textbf{Scene Awareness \& Object Understanding (1--10)}: Measures how
	well the response interprets objects, their positions, and actions.
	9--10: Clearly understands all relevant objects and their relationships.
	7--8: Minor misinterpretations but mostly correct.
	5--6: Some key objects misunderstood or ignored.
	3--4: Many errors in object recognition and reasoning.
	1--2: Misidentifies or ignores key objects.
	
	\item \textbf{Repetition-Token (1--10)}: Identifies unnecessary repetition in
	reasoning.
	9--10: No redundancy, very concise.
	7--8: Minor repetition but still clear.
	5--6: Noticeable redundancy.
	3--4: Frequent repetition that disrupts reasoning.
	1--2: Excessive redundancy, making reasoning unclear.
	
	\item \textbf{Hallucination (1--10)}: Detects irrelevant or invented reasoning
	steps not aligned with ground truth.
	9--10: No hallucinations, all reasoning is grounded.
	7--8: One or two minor hallucinations.
	5--6: Some fabricated details.
	3--4: Frequent hallucinations.
	1--2: Majority of reasoning is hallucinated.
	
	\item \textbf{Semantic Coverage-Step (1--10)}: Checks if the response fully
	covers the critical reasoning elements.
	9--10: Nearly complete semantic coverage.
	7--8: Good coverage, some minor omissions.
	5--6: Partial coverage with key gaps.
	3--4: Major gaps in reasoning.
	1--2: Very poor semantic coverage.
	
	\item \textbf{Commonsense Reasoning (1--10)}: Assesses the use of intuitive
	driving logic in reasoning.
	9--10: Displays strong commonsense understanding.
	7--8: Mostly correct, with minor gaps.
	5--6: Some commonsense errors.
	3--4: Frequent commonsense mistakes.
	1--2: Lacks basic driving commonsense.
	
	\item \textbf{Missing Step (1--10)}: Evaluates if any necessary reasoning steps
	are missing.
	9--10: No critical steps missing.
	7--8: Minor missing steps, but answer is mostly intact.
	5--6: Some important steps missing.
	3--4: Many critical reasoning gaps.
	1--2: Response is highly incomplete.
	
	\item \textbf{Relevance (1--10)}: Measures how well the response is specific to
	the given scenario and ground truth.
	9--10: Highly specific and directly relevant to the driving scenario.
	7--8: Mostly relevant, but some minor parts may be overly generic.
	5--6: Somewhat relevant but lacks precision; contains vague or general reasoning.
	3--4: Mostly generic or off-topic reasoning, with significant irrelevant content.
	1--2: Largely irrelevant, missing key aspects of the scenario.
	
	\item \textbf{Missing Details (1--10)}: Evaluates the extent to which critical
	information is missing from the response, impacting the reasoning quality.
	9--10: No significant details are missing; response is comprehensive.
	7--8: Covers most important details, with minor omissions.
	5--6: Some essential details are missing.
	3--4: Many critical reasoning steps or contextual details are absent.
	1--2: Response is highly lacking in necessary details.
\end{enumerate}

\emph{Final Evaluation: Compute the Overall Score as the average of all metric
	scores. Avoid subjective interpretation and adhere to the given thresholds.
	Always strictly follow these scoring guidelines. Do not add any additional
	explanations beyond the structured JSON output.}

\paragraph{Output format.}
The judge returns a flat JSON object with one numeric field per metric plus an
\texttt{Overall Score}, enforced by a strict \texttt{json\_schema} response
format so that unparseable or partial responses are rejected at the API level:
\begin{quote}\small\ttfamily
	\{"Faithfulness-Step": 6.0, "Informativeness-Step": 6.5,\\
	"Risk Assessment Accuracy": 7.0, "Traffic Rule\\
	Adherence": 7.5, "Scene Awareness \& Object\\
	Understanding": 8.0, "Repetition-Token": 7.0,\\
	"Hallucination": 8.5, "Semantic Coverage-Step": 7.5,\\
	"Commonsense Reasoning": 7.0, "Missing Step": 8.5,\\
	"Relevance": 8.5, "Missing Details": 7.0,\\
	"Overall Score": 7.42\}
\end{quote}

\paragraph{Two notes on the reference implementation.}
First, the reference script contains a naming inconsistency: the enforced JSON
schema declares the ninth field as \texttt{Commonsense}, whereas the system
prompt and the in-prompt example both name it \texttt{Commonsense Reasoning}.
We preserve the prompt's naming in all released per-question outputs. Second,
as stated in Appendix~\ref{app:metrics}, we compute the reported reasoning score
as the mean of the twelve dimensions rather than reading the judge's
self-reported \texttt{Overall Score} field, because the latter is occasionally
inconsistent with the dimensions returned in the same call.

\section{Additional Experimental Results}
\label{app:add_results}

\subsection{Tool-Use Behavior and Efficiency}
\label{app:tool_eff}
Table~\ref{tab:eff_component} shows that end-to-end latency is dominated by LLM
decoding (median $34.96$\,s), while tool execution adds only $5.4$\,ms at the
median per query. Per-call tool latency stays in the millisecond range on both
benchmarks (Table~\ref{tab:eff_benchmark}); the lone $35.70$\,s maximum comes
from a single cold-start monocular-depth call, invoked in only $26$ of all
queries (Table~\ref{tab:eff_tool}). Inference peaks at $\sim$19\,GB on
DriveLMM-o1 and $\sim$24\,GB on DriveMLLM, so the full system runs on a single
consumer GPU. Overall, adding a hierarchical memory and a tool-use loop
introduces negligible runtime overhead relative to the backbone VLM's own
decoding cost.

\begin{table}
	\centering \small
	\begin{tabular}{l ccc}
		\toprule
		Component & Median & p90 & Max\\
		\midrule
		LLM generation             & 34.96\,s & 50.66\,s & 82.60\,s\\
		Tool execution$^{\dagger}$ & 5.4\,ms  & 84.1\,ms & 35.70\,s\\
		Parsing                    & 0.13\,ms & 0.22\,ms & 0.46\,ms\\
		Other (decode, tokenize)   & 49.1\,ms & 91.7\,ms & 2.08\,s\\
		\midrule
		End-to-end                 & \textbf{35.04\,s} & \textbf{50.73\,s} & \textbf{96.75\,s}\\
		\bottomrule
	\end{tabular}
	\caption{Per-query latency breakdown. $^{\dagger}$aggregated over all tool calls in a query.}
	\label{tab:eff_component}
\end{table}

\begin{table}
	\centering
	\small
	\setlength{\tabcolsep}{4pt}
	\begin{tabular}{l r cc c}
		\toprule
		Benchmark & Calls & Median & p90 & Peak mem.\\
		\midrule
		DriveLMM-o1 & 1295 & 0.20\,ms & 1.49\,ms  & $\sim$19\,GB\\
		DriveMLLM   & 1956 & 2.12\,ms & 74.15\,ms & $\sim$24\,GB\\
		\bottomrule
	\end{tabular}
	\caption{Per-benchmark tool-call latency (median/p90) and inference peak GPU memory. The rare 35.70\,s max (one cold-start monocular-depth call) is reported in Table~\ref{tab:eff_component}.}
	\label{tab:eff_benchmark}
\end{table}

\begin{table}
	\centering
	\small
	\setlength{\tabcolsep}{4pt}
	\begin{tabular}{l r cc}
		\toprule
		Tool & $N$ & Median & p90\\
		\midrule
		Instance filter (Python)     & 419 & 0.05\,ms & 0.06\,ms\\
		2D detection (YOLO-World)    & 960 & 2.08\,ms & 80.8\,ms\\
		3D localization (camera)     & 390 & 2.92\,ms & 5.04\,ms\\
		3D localization (monocular)  & 26  & 99.6\,ms & 241\,ms\\
		Geometry / distance / depth  & 119 & $\sim$5\,ms & $\sim$9\,ms\\
		\bottomrule
	\end{tabular}
	\caption{Per-tool latency. Only monocular depth (UniDepth) is costly, and it is invoked rarely ($N{=}26$); 2D detection uses per-image caching.}
	\label{tab:eff_tool}
\end{table}

\subsection{Confidence Intervals}
\label{app:stats}
The main results are single deterministic evaluation runs: inference uses greedy
decoding at temperature $0$, so repeated evaluation of a fixed checkpoint is
bit-identical and run-to-run variance is zero by construction. Sampling
variability nonetheless remains because the test split is finite. We therefore
report non-parametric confidence intervals obtained by bootstrapping over test
questions ($10{,}000$ resamples with replacement, per-question scores held
fixed).

\begin{table}
\centering
\small
\setlength{\tabcolsep}{4pt}
\begin{tabular}{lcc}
\toprule
Metric & Value & 95\% CI\\
\midrule
MCQ  & 79.09 & [77.46,\ 80.72]\\
Reasoning score     & 80.03 & [79.46,\ 80.57]\\
\midrule
Risk Assessment        & 76.95 & [76.27,\ 77.59]\\
Traffic Rule Adherence & 86.29 & [85.75,\ 86.82]\\
Scene Awareness        & 84.06 & [83.55,\ 84.56]\\
Relevance              & 77.38 & [76.66,\ 78.08]\\
Missing Details        & 76.39 & [75.67,\ 77.11]\\
Faithfulness-Step      & 74.56 & [73.99,\ 75.13]\\
Informativeness-Step   & 85.53 & [85.13,\ 85.94]\\
Repetition-Token       & 96.71 & [96.49,\ 96.92]\\
Hallucination          & 76.72 & [75.99,\ 77.42]\\
Semantic Coverage-Step & 61.26 & [60.50,\ 62.06]\\
Commonsense Reasoning  & 85.46 & [84.97,\ 85.97]\\
Missing Step           & 79.13 & [78.51,\ 79.77]\\
\bottomrule
\end{tabular}
\caption{Bootstrap $95\%$ confidence intervals on DriveLMM-o1
($10{,}000$ resamples over the $2{,}391$ MCQ or $4{,}634$ full-split questions).}
\label{tab:ci}
\end{table}

The advantage in answer accuracy is robust: the strongest published baseline
reaches $71.35$ MCQ, far outside the interval under convention. The
advantage in reasoning score is not: the strongest baseline reaches $79.68$,
which falls inside our interval $[79.46, 80.57]$. We therefore claim a clear
improvement in answer accuracy and parity in overall reasoning score. This is
consistent with the shape of Table~\ref{tab:drivelmmo1}, where our model leads on Rule
Adherence and Scene Awareness while conceding the descriptive dimensions, and
with the reward design of Appendix~\ref{app:grpo}: the reinforcement stage
optimizes answer correctness on multiple-choice items and does not directly
reward descriptive completeness. Semantic Coverage-Step ($61.26$) is our lowest
dimension by over $13$ points, reflecting that the reasoning loop stops as soon
as the evidence needed to discriminate among options has been gathered, which is
efficient for answer accuracy but leaves reference reasoning steps uncovered.

\subsection{Other Qualitative Results}
\label{app:cases}
\begin{figure*}[t]
	\centering
	\includegraphics[width=\textwidth]{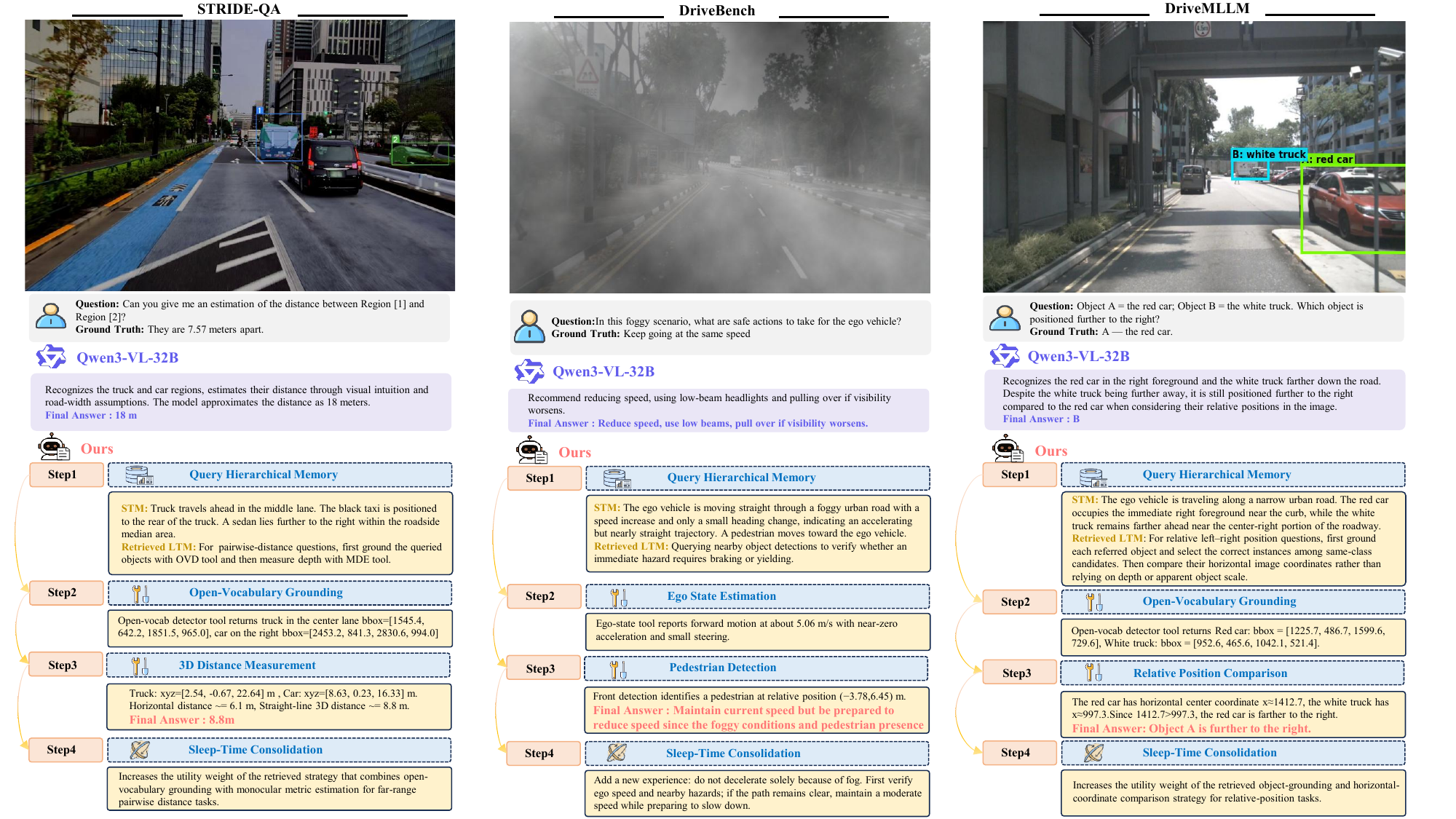}
	\caption{Additional qualitative comparison with Qwen3-VL-32B on STRIDE-QA,
		DriveBench, and DriveMLLM. Each case shows the retrieved short- and long-term
		memory, the executed tool actions with their returned observations, the final
		answer, and the offline consolidation update.}
	\label{fig:qual_supp1}
\end{figure*}

\begin{figure*}[t]
	\centering
	\includegraphics[width=\textwidth]{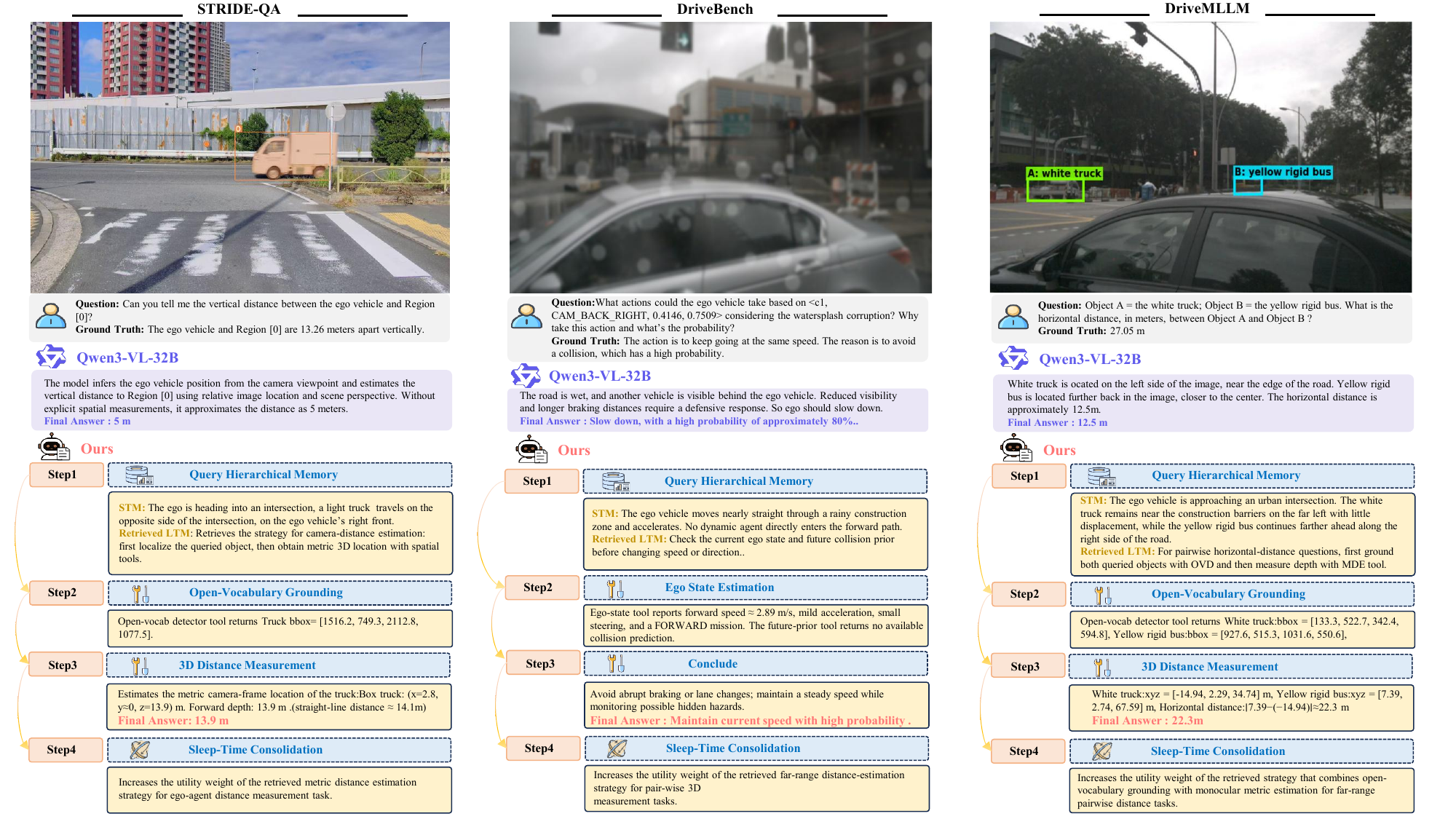}
	\caption{Further qualitative comparison in the same format.}
	\label{fig:qual_supp2}
\end{figure*}

Figures~\ref{fig:qual_supp1} and~\ref{fig:qual_supp2} extend this comparison
with six further cases. On STRIDE-QA, the baseline estimates metric distances
from image perspective and returns $5$\,m and $18$\,m for references of
$13.26$\,m and $7.57$\,m, whereas grounding the referent and querying
camera-frame 3D localization gives $13.9$\,m and $8.8$\,m. The DriveMLLM cases
show that the retrieved entry constrains which evidence is admissible rather
than merely whether a tool is called: a pairwise-distance question retrieves a
ground-then-measure procedure and recovers $22.3$\,m against a $27.05$\,m
reference, while a relative-position question retrieves a strategy that
explicitly rules out depth reasoning, and comparing horizontal image coordinates
($1412.7$ against $997.3$) identifies the correct object where the baseline's
depth-based reasoning inverts it. Under the corrupted DriveBench scenes the
baseline reacts to degraded imagery alone, recommending deceleration under
watersplash and pulling over under fog, whereas our method verifies the ego
state ($2.89$\,m/s, no collision prior) and the forward corridor before
maintaining speed, and in the fog case tempers that decision after detecting a
pedestrian at $(-3.78, 6.45)$\,m. Each trace closes with an offline
consolidation step: five cases reinforce the utility weight of the retrieved
strategy, while the fog case abstracts a new entry stating that fog alone should
not trigger deceleration before ego speed and nearby hazards are verified.
Together these cases show retrieved experience selecting task-appropriate
evidence and tool observations grounding the answer in geometric and state
evidence rather than appearance-based guesses, although the metric estimates
remain approximate.

\section{Limitations}
\label{app:limitations}

End-to-end median latency is approximately $35$\,s per query, dominated by
language-model decoding rather than tool execution; the framework is therefore
suited to offline scene understanding, data annotation, and simulation-based
evaluation rather than to a real-time driving stack, and we use driving
benchmarks as a demanding testbed for grounded multi-step reasoning rather than
as a claim of vehicle readiness. As analysed in Appendix~\ref{app:grpo}, the
reinforcement stage optimizes multiple-choice correctness, so descriptive
quality improves through supervised fine-tuning but receives no direct
reinforcement signal. Inference is deterministic, but the pipeline was trained
once: the intervals in Table~\ref{tab:ci} capture test-set sampling variability,
not variability across independent training runs, and the reinforcement stage is
not bit-reproducible even under a fixed trainer seed. Finally, offline
consolidation improves accuracy for nine generations and then declines,
indicating saturation of the store; we observe this on a single consolidation
trajectory and do not characterize its variance or asymptote. Improving
retrieval reliability and belief-conflict resolution are the natural next steps.

\end{document}